\documentclass[letterpaper, 10 pt, conference]{ieeeconf}  

\IEEEoverridecommandlockouts                              

\usepackage[pdftex]{graphicx}
\usepackage{amsmath} 
\usepackage{lipsum}
\usepackage{subcaption}
\usepackage{amssymb}
\usepackage{xcolor}
\usepackage[colorlinks=true, urlcolor=blue, linkcolor=red]{hyperref}
\usepackage{ulem}
\usepackage[most]{tcolorbox}
\usepackage{multirow}
\usepackage{gensymb}
\usepackage{array}
\usepackage{ragged2e}
\usepackage{caption}
\usepackage{adjustbox}

\title{\LARGE \bf
NeRF-enabled Analysis-Through-Synthesis for ISAR Imaging of Small Everyday Objects with Sparse and Noisy UWB Radar Data}

\author{Md Farhan Tasnim Oshim$^{1}$,  Albert Reed$^{2}$, Suren Jayasuriya$^{3}$, and Tauhidur Rahman$^{4}$
\thanks{*This work was partially supported by ONR grant N00014-23-1-2406, NSF CCF-2326905, Raytheon, Inc., Qualcomm, Google, Halıcıoğlu Data Science Institute – UC San Diego, and Manning College of Information \& Computer Sciences – UMass Amherst.}
\thanks{$^{1}$Md Farhan Tasnim Oshim is with Manning College of Information and Computer Sciences, University of Massachusetts Amherst, MA, USA
        {\tt\small farhanoshim@cs.umass.edu}}%
\thanks{$^{2}$Albert Reed with the Kitware INC, NC, USA
        {\tt\small albertnm123@gmail.com}}%
\thanks{$^{3}$Suren Jayasuriya is with the Schools of Arts, Media and Engineering and Electrical, Computer and Energy Engineering, Arizona State University, Tempe, AZ, USA
        {\tt\small sjayasur@asu.edu}}%
\thanks{$^{4}$Tauhidur Rahman is with Halıcıoğlu Data Science Institute,
University of California San Diego, CA, USA
        {\tt\small trahman@ucsd.edu}}%
}

\begin{document}

\maketitle
\thispagestyle{empty}
\pagestyle{empty}

\begin{abstract}

Inverse Synthetic Aperture Radar (ISAR) imaging presents a formidable challenge when it comes to small everyday objects due to their limited Radar Cross-Section (RCS) and the inherent resolution constraints of radar systems. Existing ISAR reconstruction methods including backprojection (BP) often require complex setups and controlled environments, rendering them impractical for many real-world noisy scenarios. In this paper, we propose a novel Analysis-through-Synthesis (ATS) framework enabled by Neural Radiance Fields (NeRF) for high-resolution coherent ISAR imaging of small objects using sparse and noisy Ultra-Wideband (UWB) radar data with an inexpensive and portable setup. Our end-to-end framework integrates ultra-wideband radar wave propagation, reflection characteristics, and scene priors, enabling efficient 2D scene reconstruction without the need for costly anechoic chambers or complex measurement test beds. With qualitative and quantitative comparisons, we demonstrate that the proposed method outperforms traditional techniques and generates ISAR images of complex scenes with multiple targets and complex structures in Non-Line-of-Sight (NLOS) and noisy scenarios, particularly with limited number of views and sparse UWB radar scans. This work represents a significant step towards practical, cost-effective ISAR imaging of small everyday objects, with broad implications for robotics and mobile sensing applications.

\end{abstract}

\section{Introduction}

Inverse Synthetic Aperture Radar (ISAR) imaging is a standard radar mode employed for target identification, typically involving a stationary radar and a maneuvering target to collect various aspects of target reflectivity through motion. To achieve a high-resolution image of the target in the scene, reflected signals are coherently accumulated to obtain a sufficiently large synthetic aperture. Although Synthetic Aperture Radar (SAR) imaging of large objects such as UAVs, drones, vessels, buildings, and cities is widespread \cite{Chenchen2016drones, li2018wide, BrennercitySAR}, imaging of small targets remains challenging due to their small Radar Cross-Section (RCS) and limitations in radar range resolution. Existing ISAR image reconstruction methods typically require complex hardware setups \cite{zhao2022}, high-precision measurement test beds \cite{yanik_testbed}, or noise-less expensive anechoic chambers \cite{Sakamoto2015, Kang2022}. We propose an analysis-through-synthesis optimization framework that leverages recent advances in neural rendering \cite{mildenhall2021nerf} to perform high-resolution radar imaging that works in diverse conditions employing merely an inexpensive, easy-to-use, portable setup. The proposed framework integrates the physics governing radar signal formation, scene priors, and noise models to enable coherent ISAR image reconstructions.

The Neural Radiance Field (NeRF) technique \cite{mildenhall2021nerf} employs neural networks and differentiable volume rendering to generate new perspectives of 3D scenes from 2D images. Since its inception, NeRF has spurred substantial research efforts \cite{martin2021nerf, RegNeRF}, resulting in optimizations and expansions that improve accuracy and efficiency in synthesizing novel views and reconstructing 3D geometry from measurements. Recently, these techniques have extended to other sensing modalities such as sonar \cite{reed2023neural, reed2021implicit, reed2022sinr}, radar \cite{CoIR, RaNeRF, ISAR_NeRF, CSAR_Incorherent}, and lidar \cite{INF_LiDar, MOISST, zhong2023shine}. Reconstructing large targets such as satellites, buildings, or city maps has been explored using NeRF for these modalities. While RaNeRF \cite{RaNeRF} and ISAR-NeRF \cite{ISAR_NeRF} focus on neural rendering for space target reconstruction,  CSAR-Incoherent \cite{CSAR_Incorherent} and MOISST \cite{MOISST} reconstruct urban scenes. However, imaging small targets has always posed a challenge due to their minimal Radar Cross-Section (RCS) and sensor resolution limitations. Recent work of Reed et al. \cite{reed2023neural} tackles small target synthesis with sonar but faces challenges with imaging Non-Line-of-Sight (NLOS) scenes due to the acoustic signal's inability to penetrate through different materials.

We present a novel 2D ISAR reconstruction algorithm, utilizing an analysis-through-synthesis optimization approach enabled by conventional NeRF \cite{mildenhall2021nerf} techniques, yet diverging notably in sampling (spherical instead of line sampling) and output representation (time series rather than intensity). Through numerous experiments on simulated and hardware-measured data, we demonstrate both quantitative and qualitative superiority of our approach over traditional techniques. The key contributions of our work can be summarized as -

\begin{itemize}
    \item We introduce an end-to-end analysis-through-synthesis (ATS) framework that integrates ultra-wideband radar wave propagation and reflection characteristics to facilitate 2D scene reconstruction.
    \item Our approach generates ISAR images without any costly anechoic chambers or complex measurement test beds, thereby reducing both cost and computation time in the reconstruction process.
    \item Our method demonstrates superior performance compared to conventional backprojection (BP) in both simulated and real scenes with multiple targets and complex structures in Non-Line-of-Sight (NLOS) and noisy environments, particularly with limited number of views and sparse UWB radar scans.
\end{itemize}

\begin{figure}[b!]
    \centering
    \includegraphics[width=0.4\textwidth]{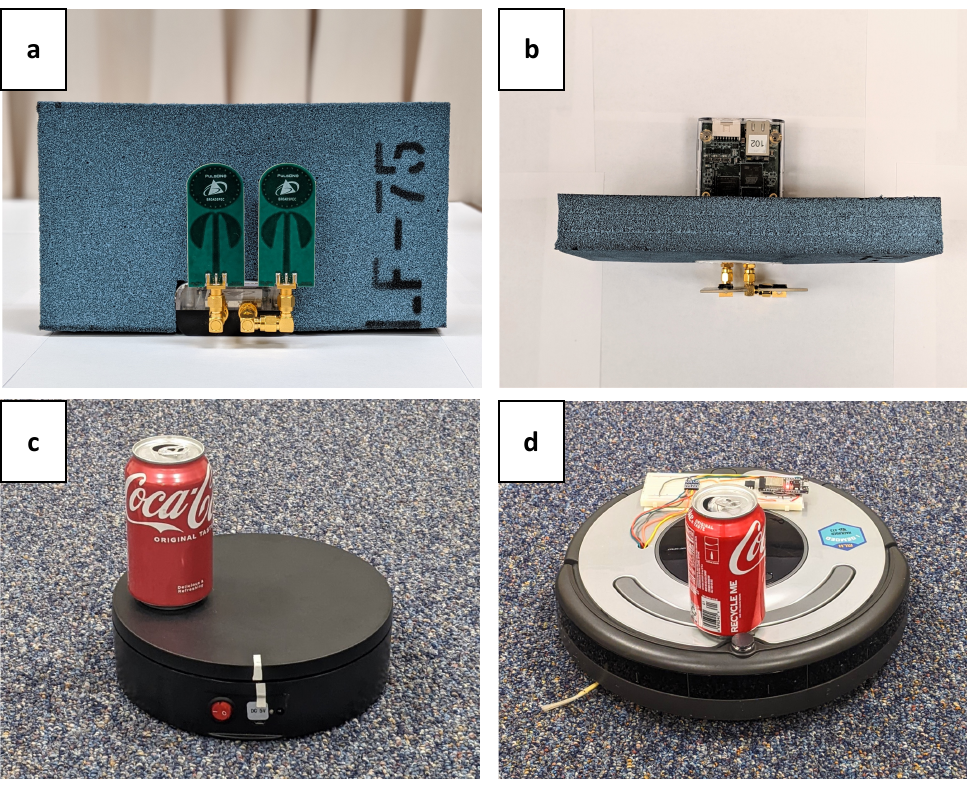} 
   
    \caption{Hardware and Measurement Setup: (a) Front view of Time Domain P440 radar module with absorber behind the antenna (b) Top view of the radar (c) JAYEGT \cite{JAYEGT} electronically motorized turn-table (d) Electronically enhanced Floor Cleaning Robot \cite{Roomba_smarter}} 
    \label{fig: hardware_setup}
\end{figure}

\begin{figure*}[t!]
    \centering
    \includegraphics[width=0.8\textwidth]{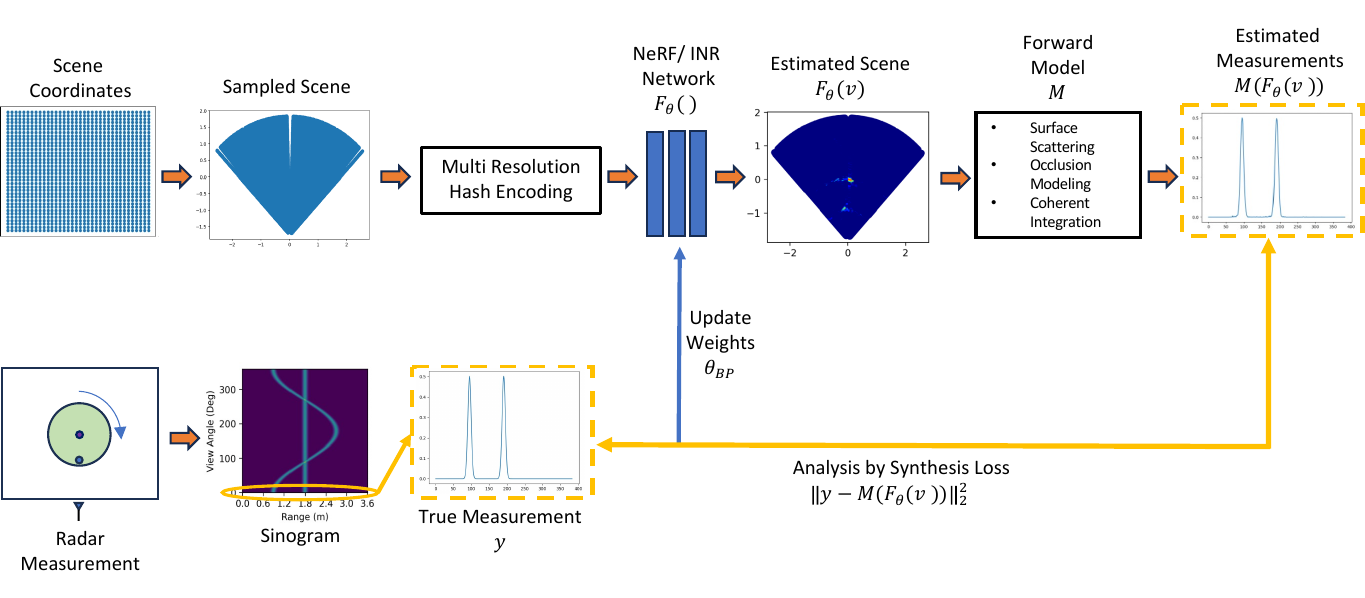} 

    \caption{illustrates our proposed analysis-through-synthesis (ATS) pipeline for radar image reconstruction. First, scene coordinates are sampled using our spherical sampling scheme and encoded with a multi-resolution hash encoding. Next, the encoded coordinates are fed into the NeRF network to predict scene scattering functions, which are then used to generate radar measurements through a differentiable forward model. Training the network involves minimizing the loss between estimated and true radar measurements for each scan of the sinogram.}
    \label{fig: AtS_pipeline}
\end{figure*}

\section{Related Works}

\subsection{ISAR Imaging}

The concept of inverse synthetic aperture radar (ISAR) emerged to resolve imaging scenarios involving moving targets and stationary radars. Circular ISAR methods, specifically designed for stationary radar systems, facilitate high-resolution imaging of rotating targets by leveraging their circular motion. In ISAR imaging, conventional reconstruction methods often necessitate intricate hardware configurations \cite{zhao2022}, precise measurement environments \cite{yanik_testbed}, or costly anechoic chambers with minimal electromagnetic interference \cite{Sakamoto2015, Kang2022}. These requirements pose challenges in terms of accessibility and resource expenditure, limiting the widespread adoption and scalability. Common approaches for ISAR imaging include time-domain backprojection \cite{frey2005study}, a technique that reconstructs images by correlating received radar data with expected echoes from target positions across the scene, the Range-Doppler technique \cite{benjamin2001range}, which processes radar data collected over multiple rotations, and the Polar Format Algorithm (PFA) \cite{zhu2008polar}, which directly transforms radar data from polar to Cartesian coordinates for wavefront curvature correction to compensate for distortions in received signals. However, these algorithms often face challenges of non-uniform motion, cluttered environments, limited measurement coverage, noise, motion compensation errors, and limited resolution.

\subsection{Deep Learning-based Imaging}

Recent advancements in radar signal processing, coupled with the emergence of machine learning techniques, have shown promise in overcoming these limitations. Deep learning, in particular, has emerged as a powerful tool for radar image enhancement, target recognition, and clutter suppression. Convolutional neural networks (CNNs) and generative adversarial networks (GANs) have demonstrated remarkable capabilities in radar image processing. For ISAR imaging, Hu et al. \cite{hu2019inverse} introduced a CNN-based approach for reconstructing high-resolution images from radar measurements. Additionally, Tan et al. \cite{tan2021cnn} demonstrated a CNN-based method for SAR image denoising, showcasing the potential of CNNs in improving SAR image quality and utility. GANs, on the other hand, have been used for radar image enhancement, exploiting their ability to generate realistic and high-fidelity reconstructions from noisy or degraded input \cite{li2022unblurring, wang2018super}.

\subsection{NeRF-based Techniques}

The Neural Radiance Fields (NeRF) framework \cite{mildenhall2021nerf}, initially introduced for synthesizing photorealistic 3D scenes from images, presents a fascinating approach to improve ISAR imaging capabilities. Capitalizing on NeRF's ability to generate novel views, researchers have adopted this approach to tackle challenges in ISAR imaging including resolution enhancement, motion compensation, and clutter suppression exemplified in CoIR \cite{CoIR}, RaNeRF \cite{RaNeRF}, ISAR-NeRF \cite{ISAR_NeRF}, and CSAR-Incoherent \cite{CSAR_Incorherent} works. These techniques have explored reconstructing large targets like satellites, buildings, or even city maps using NeRF. Notably, RaNeRF \cite{RaNeRF} and ISAR-NeRF \cite{ISAR_NeRF} focus on neural rendering for space target reconstruction, while CoIR \cite{CoIR}, CSAR-Incorherent \cite{CSAR_Incorherent}, and MOISST \cite{MOISST} target the reconstruction of outdoor and urban scenes. However, these approaches mostly improve on already reconstructed SAR images for novel view synthesis rather than incorporating radar measurements as input signals themselves. Our approach, on the other hand, employs an analysis-through-synthesis optimization method, using an implicit neural representation akin to NeRF, to predict point scatterers within the scene and synthesize radar measurements through a differentiable forward model, optimizing the network by minimizing the loss between synthesized and actual data.

\section{Hardware and Measurement Setup}
We utilize a monostatic time domain Ultra-WideBand (UWB) Impulse Radar P440 \cite{P440} with time windowing capabilities for sensing, allowing us to exclude unwanted signal reflections and conduct indoor measurements without costly anechoic chambers. It operates from 3.1 to 4.8 GHz frequency centering at 4.3 GHz. Figure \ref{fig: hardware_setup} illustrates the radar hardware and the measurement setup. Our ISAR imaging measurement in an ordinary room employs the UWB radar positioned at a fixed location, using coherent pulse integration to enhance the signal-to-noise ratio, with the radar sampling rate exceeding the maximum Doppler extent for coherent processing into unaliased ISAR image. The procedure involves placing the object of interest on a rotating table in front of the radar system. We leverage JAYEGT \cite{JAYEGT}, an electronically motorized turn-table for placing the target object and getting a steady rotational motion. Additionally, we demonstrate the feasibility of collecting data with a standard floor cleaning robot by integrating supplementary circuitry (costing approximately \$10) \cite{Roomba_smarter}, which incorporates an ESP32 chip with Bluetooth functionality.

In this setting, radar measurements are acquired from $n$ virtual radar positions, or in our case, from $360\degree$ positions creating a circular synthetic aperture. These radar measurements are aggregated to create a 2D sinogram, where the x-axis represents the range in meters, and the y-axis corresponds to $\phi$ in degrees. Each $\phi$ value aligns with a radar position or a $1\degree$ rotation angle of the turn-table. While the measurements are intended to be static for each angle, in practice, data is continuously collected until the rotating table completes a full revolution. The only requirements are to maintain a constant rotational speed of the target during the sensing period and to keep the axis of rotation unchanged throughout the observation period.

\section{Analysis-Through-Synthesis (ATS) Pipeline}
After acquiring the measurements (i.e. sinograms), our reconstruction process employs an analysis-through-synthesis optimization technique, leveraging an implicit neural representation (INR) similar to NeRF, in traditional view synthesis. Figure \ref{fig: AtS_pipeline} shows our proposed analysis-through-synthesis (ATS) pipeline. At first, the scene coordinates are sampled using our spherical sampling scheme for a given sensor position. Then the sampled scene is passed through a multi-resolution hash encoding block similar to Instant-NGP \cite{muller2022instant} for generating the positional embedding $v$ from each coordinate. It uses hash functions to generate varied positional embedding, capturing relative positions in a sequence at different resolutions. In contrast to positional encoding, the multi-resolution approach allows a more dynamic and flexible representation of positional information. Then the encoded coordinates are fed into the NeRF neural network of four fully connected layers, referred to as the neural backprojection network $\mathcal{N}_{BP}$, parameterized by weights $\theta_{BP}$. Through this network, we predict the complex scattering function $\sigma^{\prime}$ in the estimated scene $F_{\theta} (v)$. More precisely, the network characterizes the scatterers at each location of the scene. The scatterers within the scene are subsequently employed to generate radar measurements (each scan in the sinogram) through our differentiable forward model described in section \ref{sec:forward_model}. Training the network involves minimizing the loss between the synthesized measurements $M(F_{\theta} (v))$ and actual radar measurements $y$ for each scan of the sinogram. In short, by employing a neural network, we predict scene scatterers, and then by employing a differentiable forward model, we synthesize radar measurements over time.

\subsection{Radar Forward Model} \label{sec:forward_model}

\begin{figure}[htb!]
    \centering
    \includegraphics[width=0.45\textwidth]{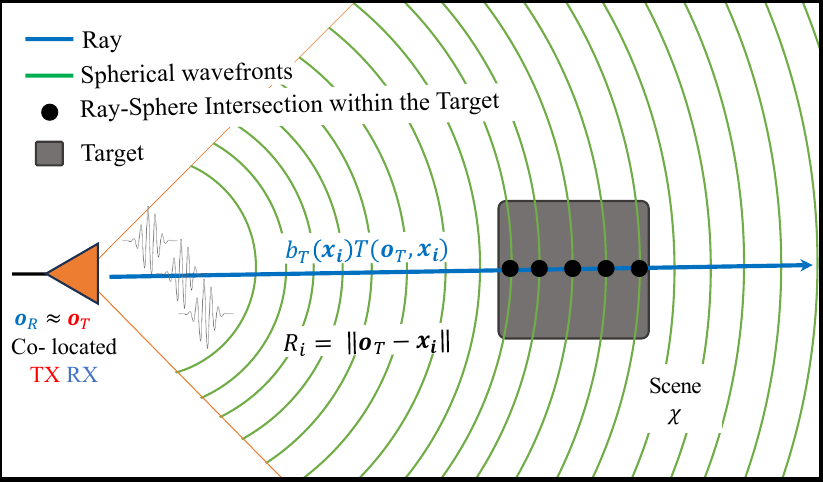} 
   
    \caption{Illustrates our forward model geometry and sampling strategy. A transmitted ray (blue) is emitted towards the scene, propagating to a scene $\chi$ weighted by the directivity function \( b_T(x) \) and transmission probability \( T(o_T, x) \). All transmitted rays within the antenna beamwidth (orange lines) are then sampled at the intersection point of the sphere (green) defined by range samples. TX and RX are collocated. }

    \label{fig: forward_model_geometry}
\end{figure}

In this subsection, we present the radar forward measurement model, which serves as the foundation for our ATS pipeline. This model draws inspiration from a point-based radar scattering model as outlined in \cite{poole2022dreamfusion, chance2022differentiable}. Such point-scattering models provide computational feasibility and differentiability, making them conducive to neural rendering.

Let $x \in \mathbb{R}^3$ describe a 3D coordinate in a scene, $\sigma(x) \in \mathbb{R}$ denote the amplitude of the scatterer at $x$ in the scene, and $s(t)$ represent the transmitted UWB signal:
\begin{equation}
    s(t) = e^{-\frac{t^2}{2\tau_0^2}} \cdot \cos\left(2\pi f_c t\right)
    \label{eqn: radar_signal}
\end{equation}

Here, \( f_c \) is the center frequency of the transmitted pulse, and \( \tau_0 \) is the standard deviation that controls the pulse width of the UWB signal.

Figure \ref{fig: forward_model_geometry} illustrates the forward model geometry, with transmitter and receiver origins designated as \( o_T \) and \( o_R \) respectively. \( b_T(x) \) represents the directivity function for the transmitter and \( T(o_T, x) \) is defined as the transmission probability between a point in the scene and the origins of the transmitter and receiver respectively, accounting for occlusion effects in our model.

Let \( R = ||o_T - x||= ||o_R - x|| \) denote the distance between the point in the scene and the origins of the transmitter and receiver. Then the receiver signals can be expressed as:
\begin{equation}
     r(t) = \int_\chi  \frac{2T(o_T, x)b_T(x)}{(4\pi)^3 R^4}  L(\sigma(x)) s\left(t - \frac{2R}{c}\right) dx 
\end{equation}

where \( \chi \) denotes the set of all coordinates in the region of interest in the scene, \( L(\sigma(x)) \) represents the Lambertian scattering model of the scene point, and \( c \) is the speed of light. Considering the co-location of transmitter and receiver antennas, we assume equal transmission and reception probabilities, hence the factor of 2 in the numerator.

However, pulses are transmitted and received within discrete radial range bins in ultra-wideband radar systems. These distances are often referred to as sets of points lying on constant time-of-flight paths, also known as radial wavefronts. Received pulses are coherently integrated based on their constant time-of-flight range bins, thereby forming radar range profiles. Consequently, we approximate the forward model similar to \cite{reed2023neural} that captures these characteristics given by the following equation:

\begin{equation}
     r\left(t=\frac{2R}{c}\right) = \int_{E_r} 2T(o_T, x)b_T(x)L(\sigma(x))dx. 
\end{equation}

Here, \( E_r \) delineates the sphere encompassing all points within the constant time of flight for the radar's forward model geometry, \( \sigma(x) \) denotes the scene scatterer at $x$, and its complex scattering function is estimated by a neural network, represented as \( \sigma(x) = \mathcal{N}_{BP}(x;\theta_{BP}) \).

\subsection{Spherical Sampling}
In this subsection, we outline our method of sampling the scene using spheres of constant time-of-flight to estimate the transmission probabilities for transmitted rays. 

In Figure \ref{fig: forward_model_geometry} each green semi-circle corresponds to a constant time-of-flight $t=2R/c$ which forms a sphere. This sphere is centered where the transmitter and receiver antennas coincide ($O_T \simeq O_R$), with a radius of $R=ct/2$. The equation representing this sphere is:
\begin{equation}
    x^2+y^2+z^2= R^2.
    \label{eqn: sphere}
\end{equation}

Now, by intersecting a sphere with radius \( \frac{ct}{2} \) with a bundle of rays originating from the transmitter and falling within the beamwidth \( \theta_{BW} \), we obtain the sampled points in the scene. The transmission ray, depicted in blue with direction \(\vec{d_{T_j}} \), is defined as 
\begin{equation}
\vec{x_{T_{ij}}} = \vec{o_T} + l_i\cdot\vec{d_{T_j}}
\end{equation}

where $l_i$ represents the depth samples along the ray calculated at which a ray intersects the sphere by substituting the ray into Equation \ref{eqn: sphere}. This substitution results in a quadratic equation from which we extract the positive root:
\begin{equation}
x = \frac{-b + \sqrt{{b^2 - 4ac}}}{2a} 
\end{equation}

where, 
 
\begin{equation}
    a = \vec{d_{T}}\cdot\vec{d_{T}},  b = 2\vec{x_{T}}\cdot\vec{d_{T}} , c = \vec{x_{T}}\cdot\vec{x_{T}} - R^2 
\end{equation}

We address occlusion by calculating transmission probabilities between the transmitter/receiver and various scene points using the method of Mildenhall et al.~\cite{mildenhall2021nerf} and Reed et al.~\cite{reed2023neural}. These probabilities are expressed as the product of exponential terms, where $|\sigma_k|$ represents the magnitude of the scattering coefficient at range bin $k$, and $|l_{k+1} - l_k|$ denotes the distance between consecutive range bins:

\[ T(o, x_{T_i}) = \prod_{k < i} e^{-\left(|\sigma_k| \cdot |l_{k+1} - l_k| \right)}.\]

Utilizing the Lambertian scattering model, the scattered intensity $L(\sigma)$ is calculated as:

\[ L(\sigma) = \sigma \cdot \frac{(x_{T_i} - o_T)}{||x_{T_i} - o_T||} \cdot 2T(o, x_{T_i}).\]

These equations facilitate the computation of transmission probabilities and scattered intensity, crucial for addressing occlusion in the scene.


\begin{figure*}[ht]
\centering
\begin{minipage}{\textwidth}
    \centering
    \scriptsize \textbf{Simulated data: Noise Free} \\
    \vspace{0.1cm}
    \begin{tabular}{m{0.5cm} m{2cm} m{2cm} m{2cm} m{2cm}}
        \parbox[c]{\hsize}{\rotatebox{90}{\centering \scriptsize \textbf{Sinogram}}} & 
        \includegraphics[scale=0.3]{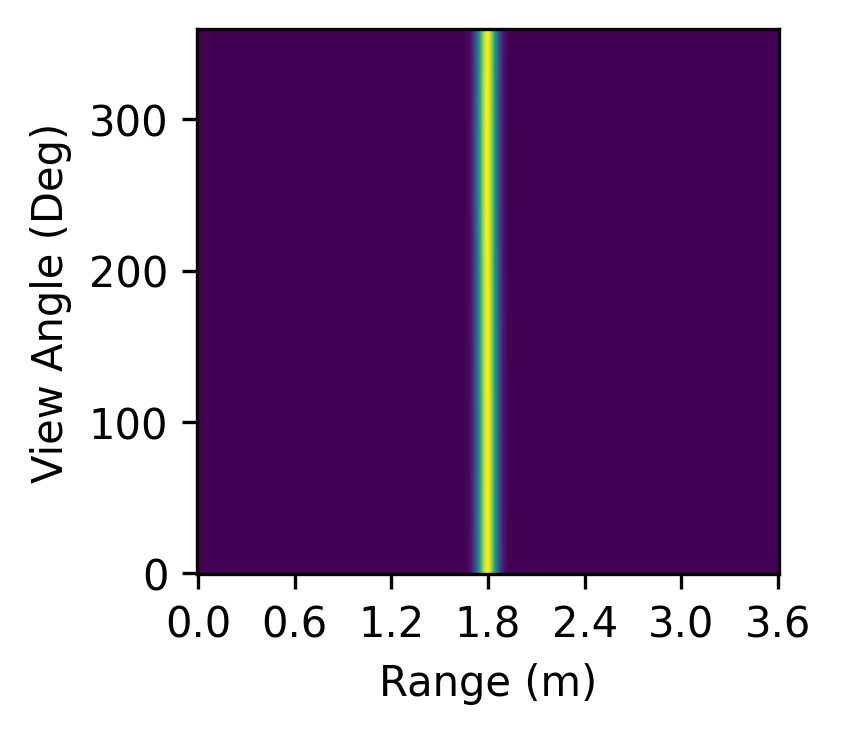} & 
        \includegraphics[scale=0.3]{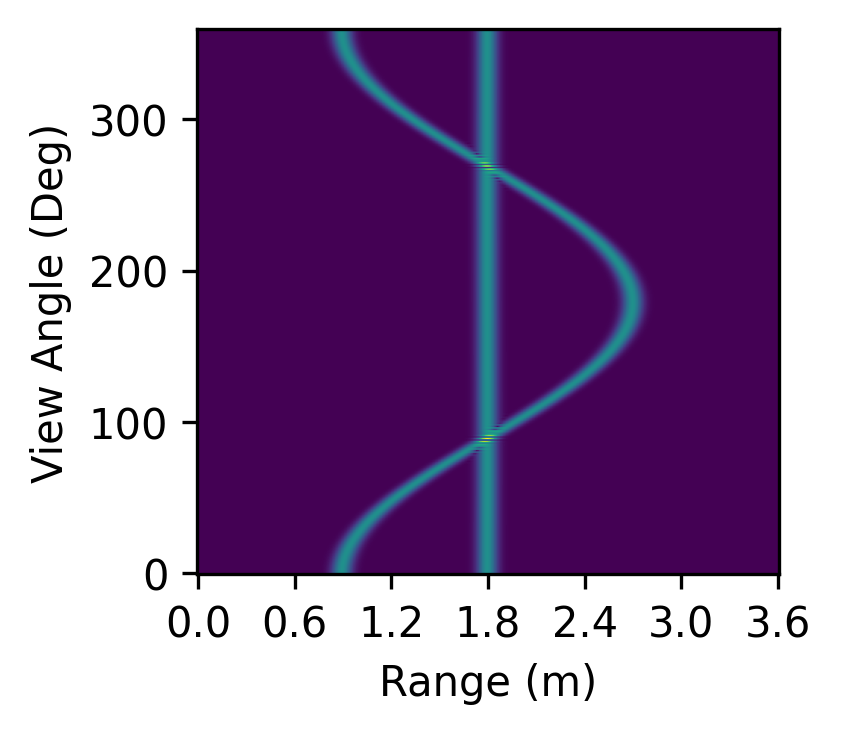} & 
        \includegraphics[scale=0.3]{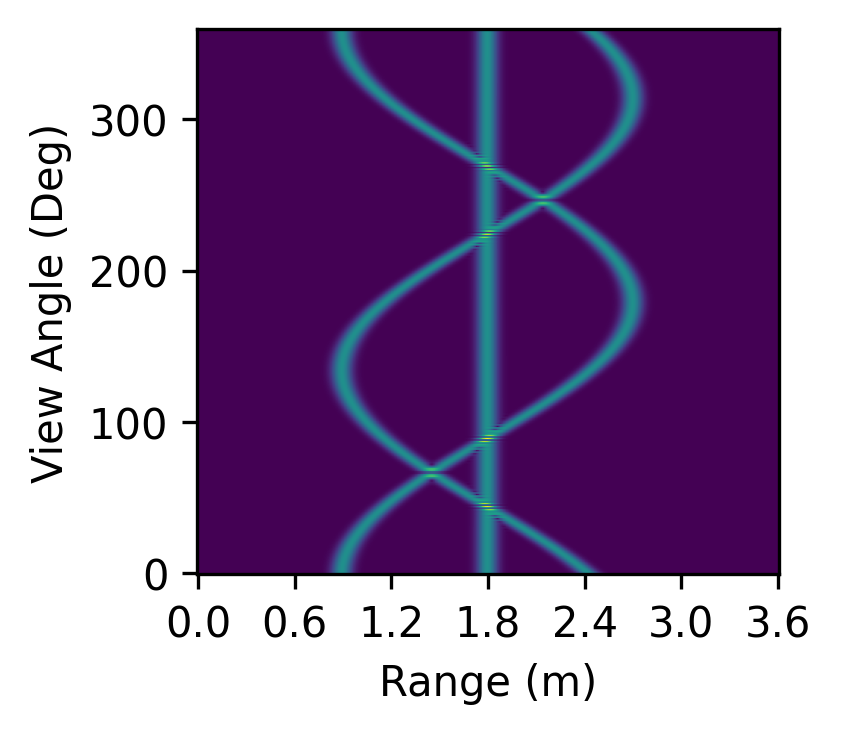} & 
        \includegraphics[scale=0.3]{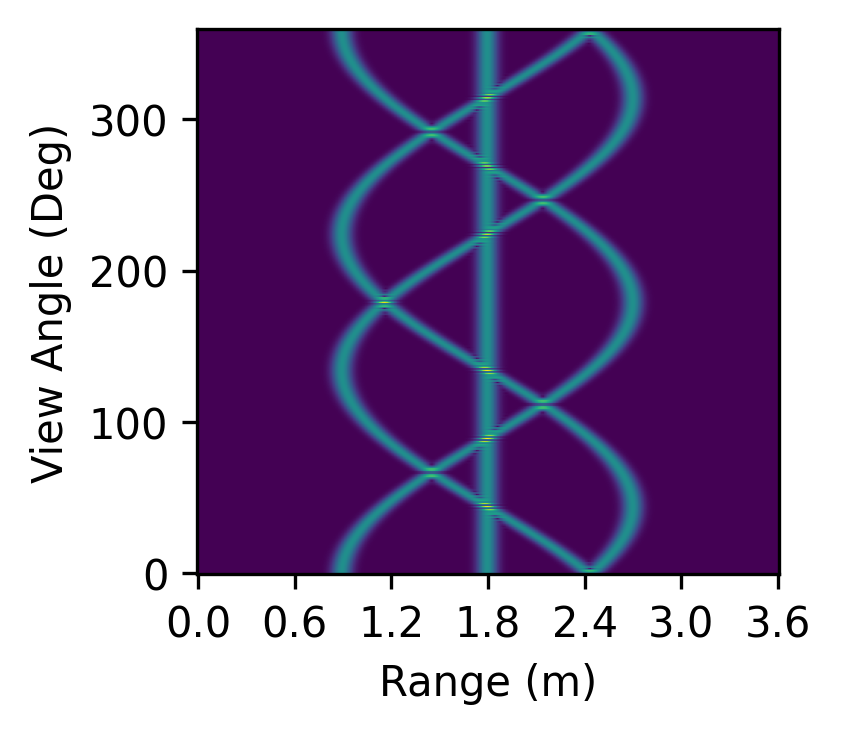} \\
        
        \parbox[c]{\hsize}{\rotatebox{90}{\centering \scriptsize \textbf{BP}}} &
        \includegraphics[scale=0.3]{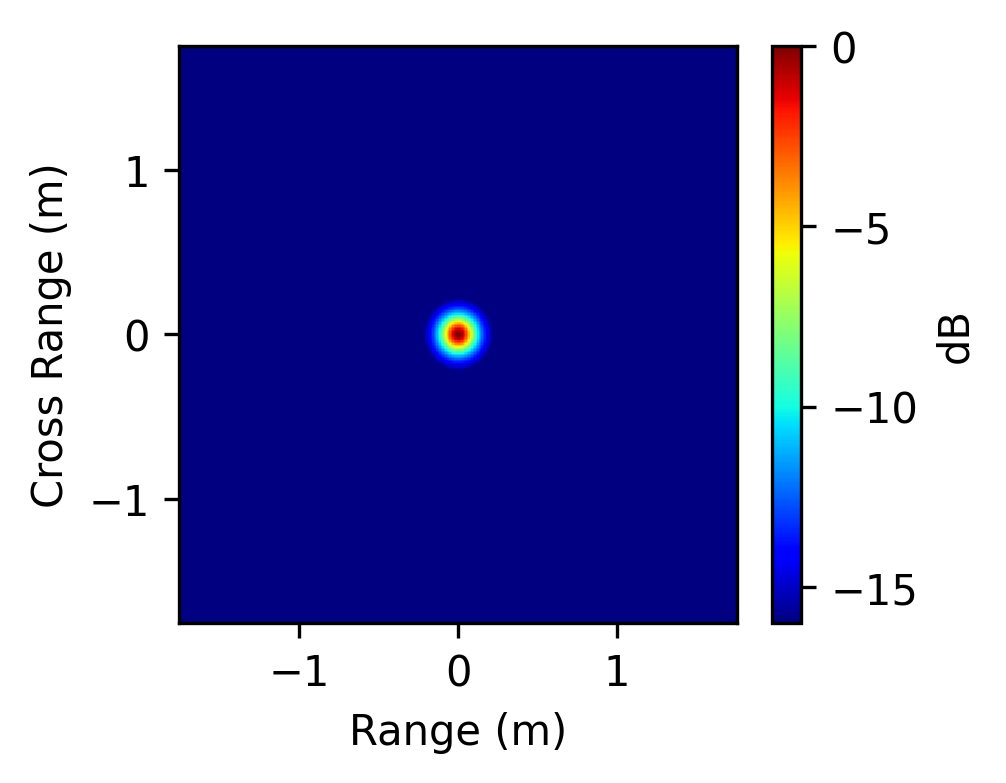} & 
        \includegraphics[scale=0.3]{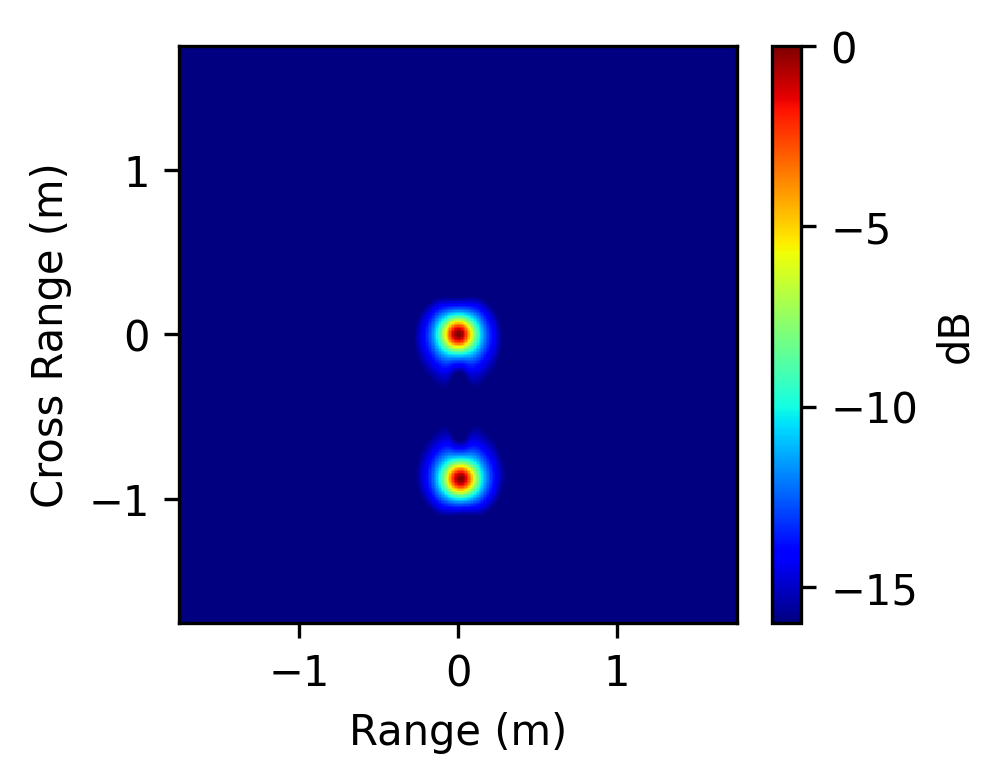} & 
        \includegraphics[scale=0.3]{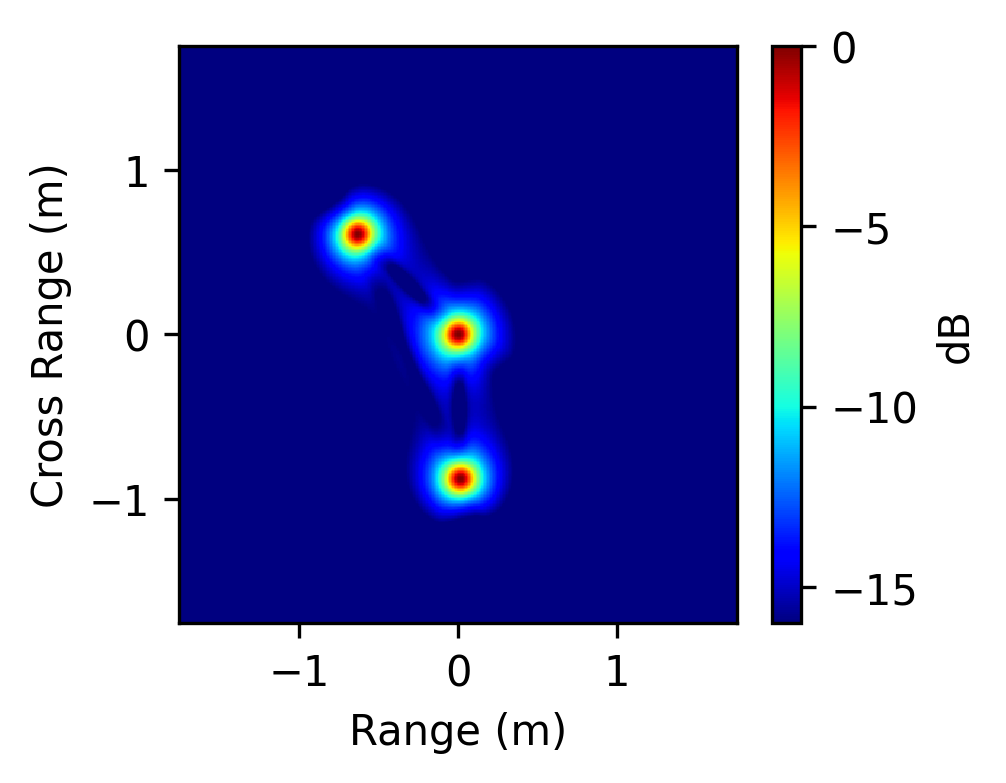} & 
        \includegraphics[scale=0.3]{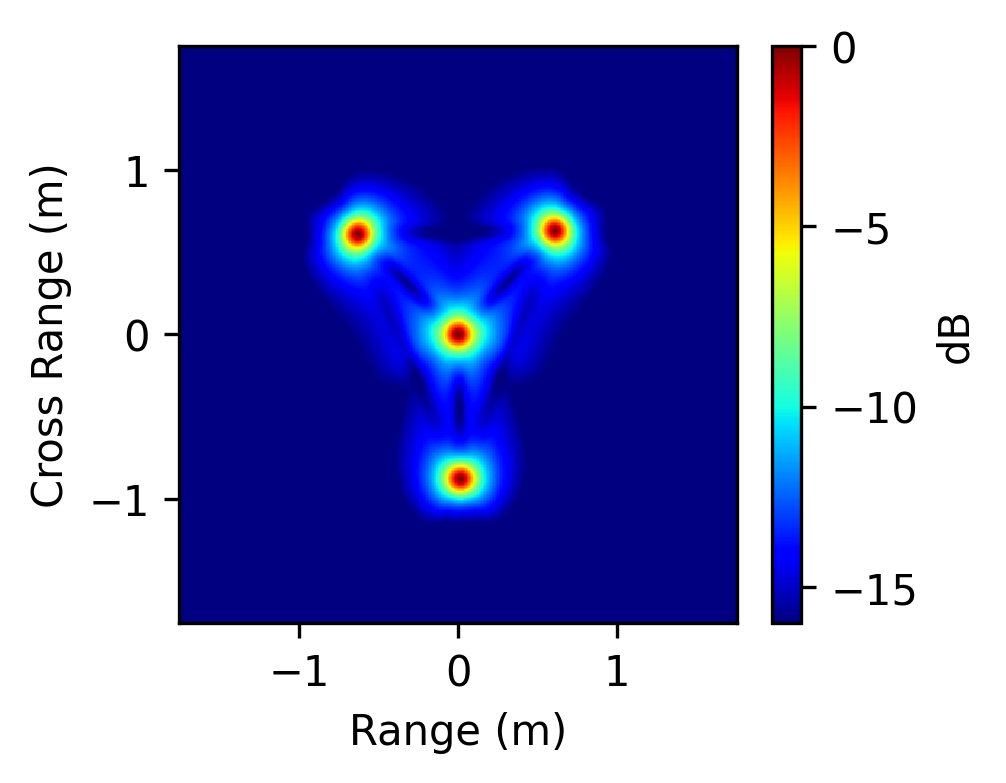} \\
        
        \parbox[c]{\hsize}{\rotatebox{90}{\centering \scriptsize \textbf{ATS}}} &
        \includegraphics[scale=0.3]{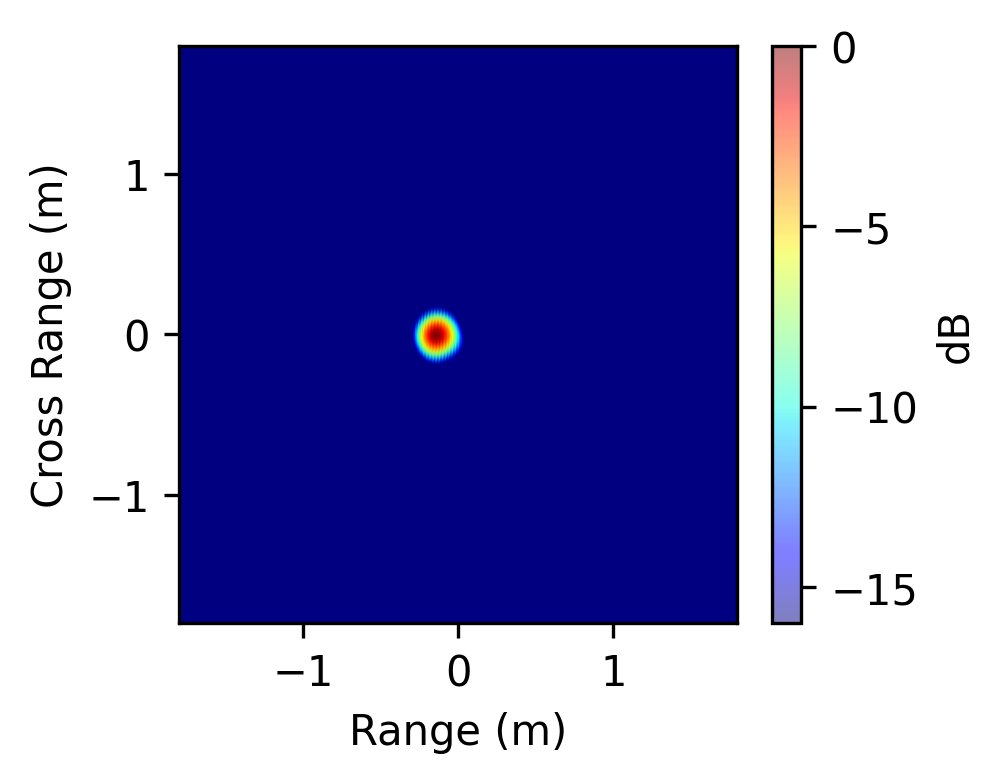} & 
        \includegraphics[scale=0.3]{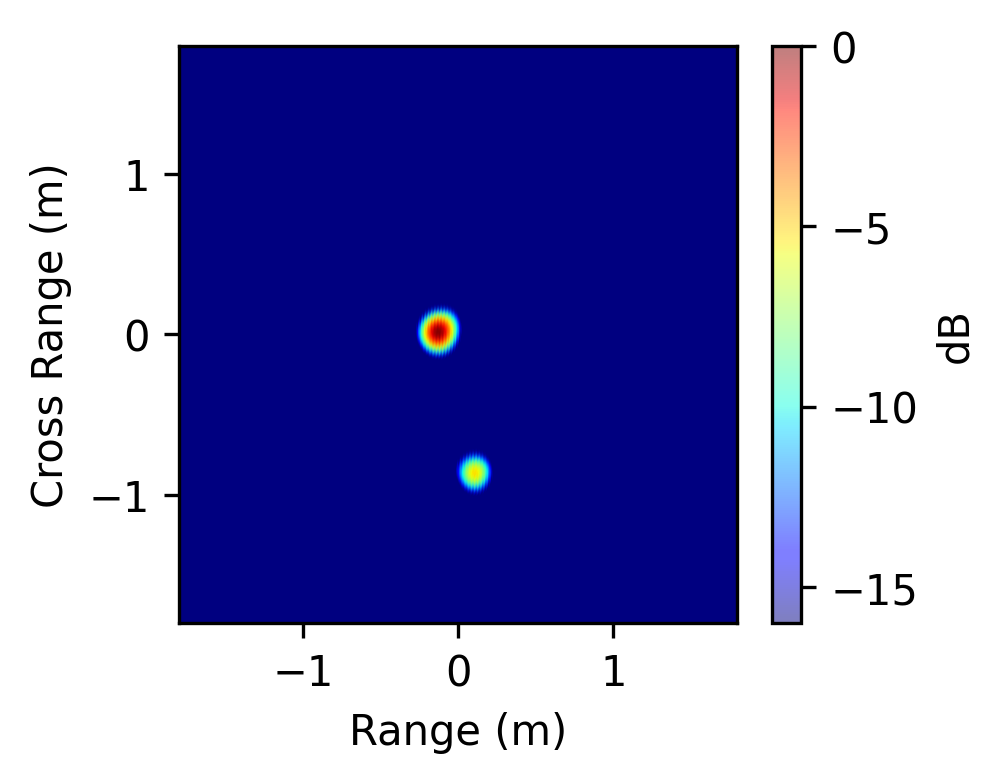} & 
        \includegraphics[scale=0.3]{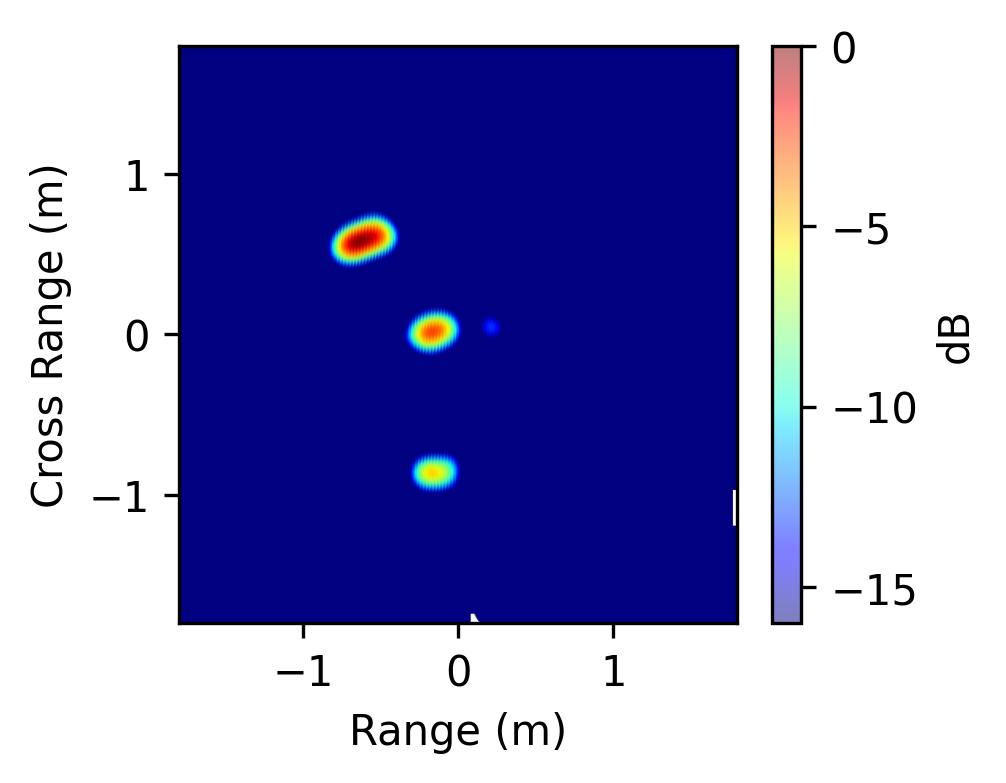} & 
        \includegraphics[scale=0.3]{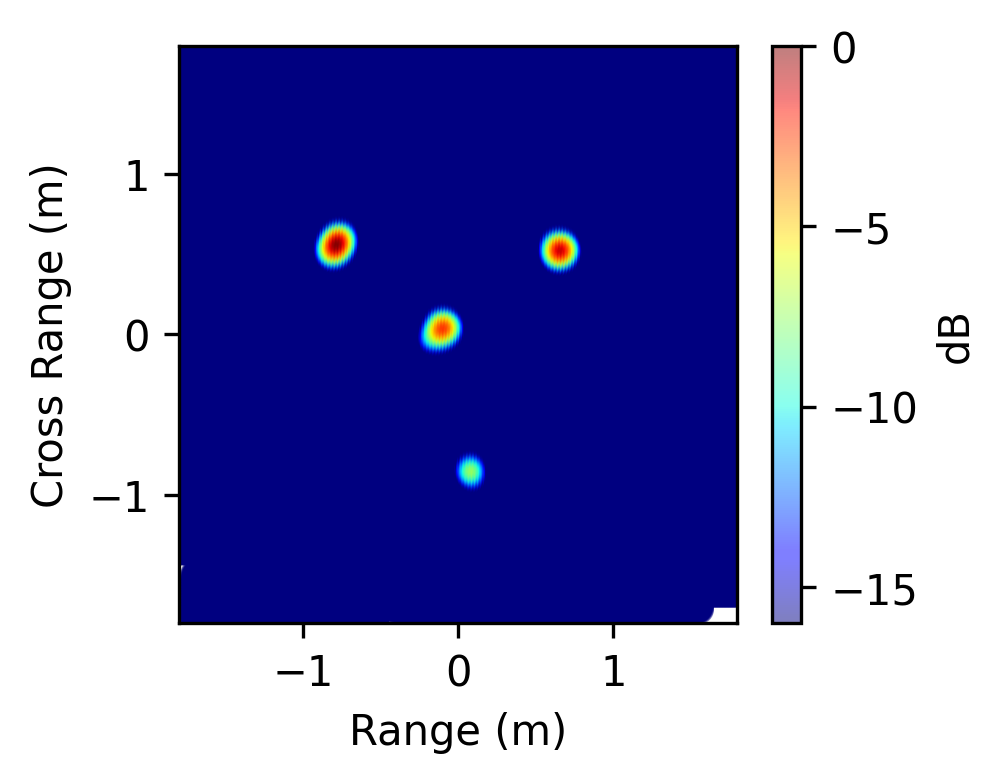} \\
    \end{tabular}
\end{minipage}

\vspace{0.3cm}

\begin{minipage}{\textwidth}
    \centering
    \scriptsize \textbf{Simulated data: With Added Gaussian Noise} \\
    \vspace{0.1cm}
    \begin{tabular}{m{0.5cm} m{2cm} m{2cm} m{2cm} m{2cm}}
        \parbox[c]{\hsize}{\rotatebox{90}{\centering \scriptsize \textbf{Sinogram}}} & 
        \includegraphics[scale=0.3]{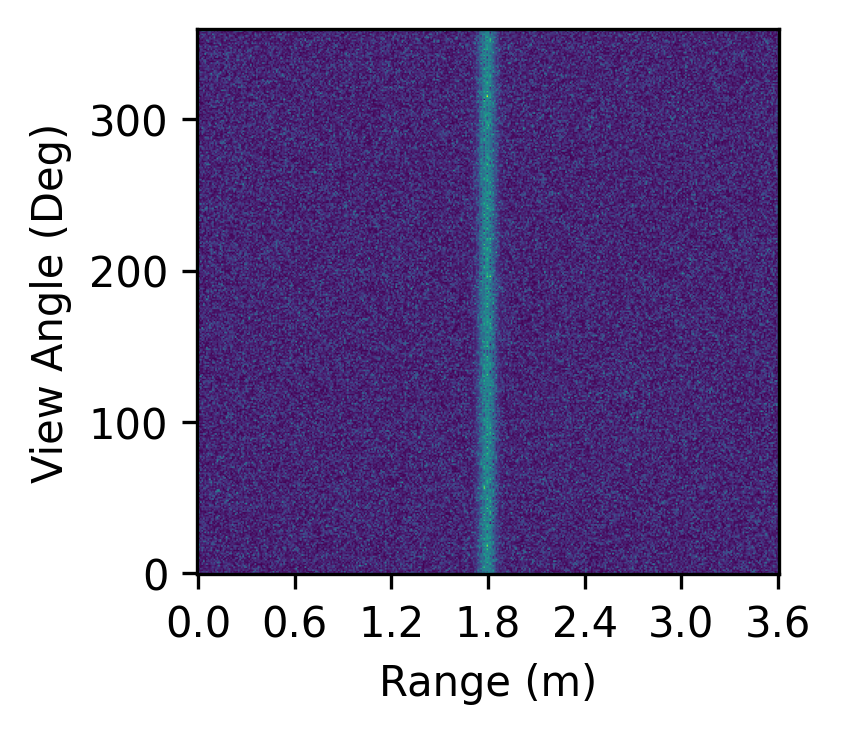} & 
        \includegraphics[scale=0.3]{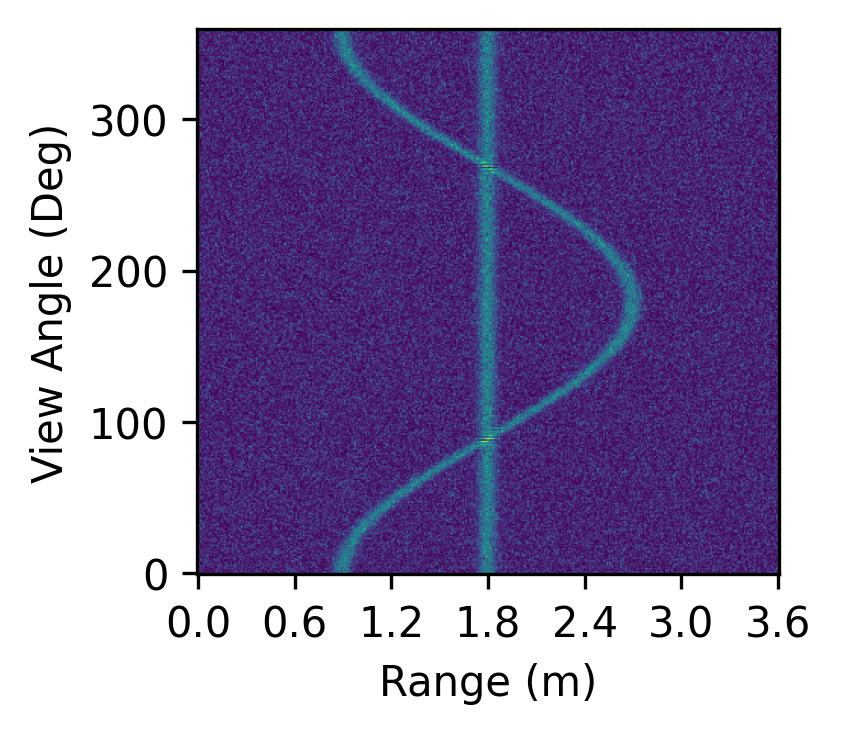} & 
        \includegraphics[scale=0.3]{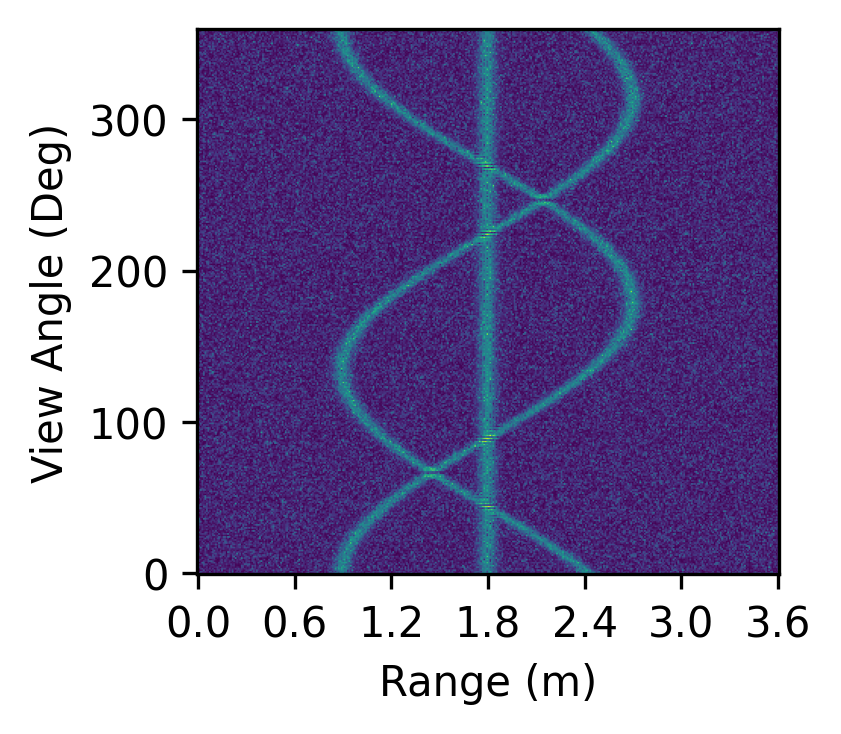} & 
        \includegraphics[scale=0.32]{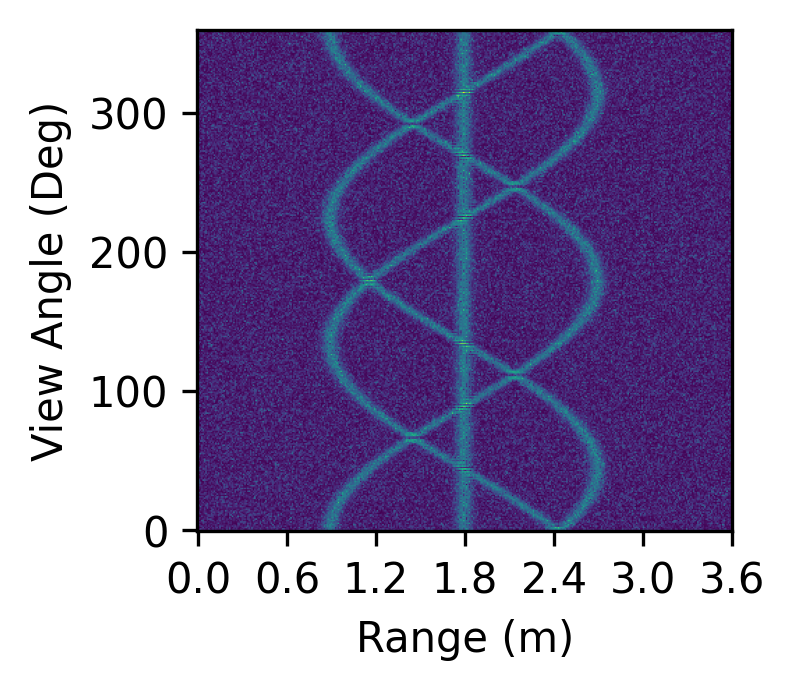} \\
        
        \parbox[c]{\hsize}{\rotatebox{90}{\centering \scriptsize \textbf{BP}}} &
        \includegraphics[scale=0.3]{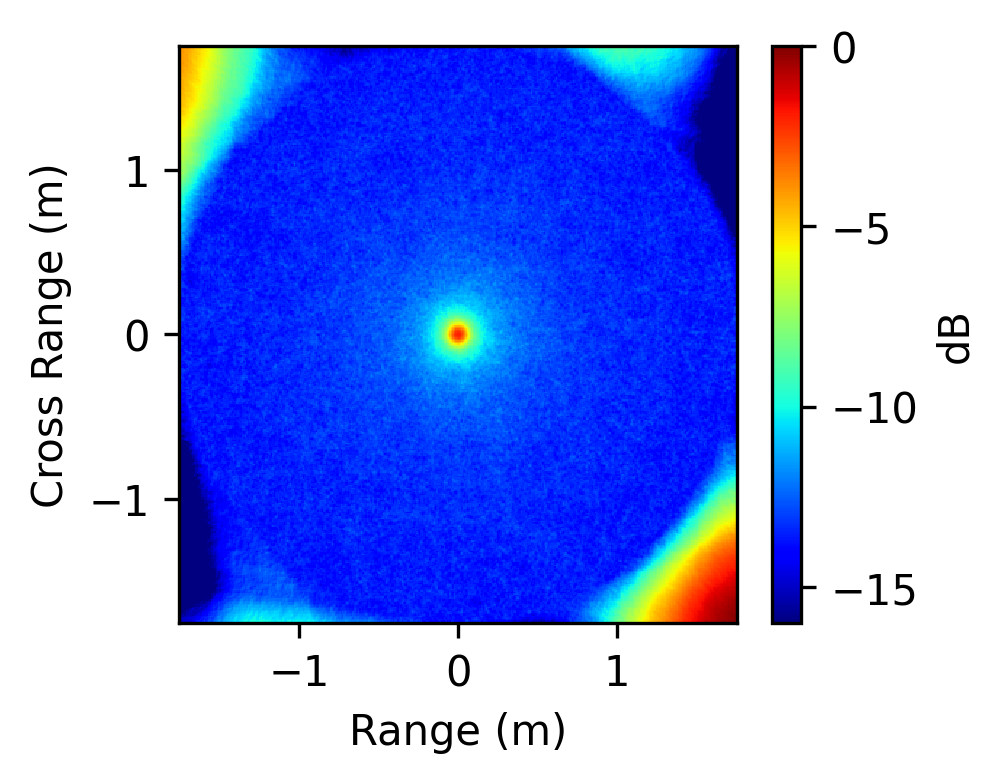} & 
        \includegraphics[scale=0.3]{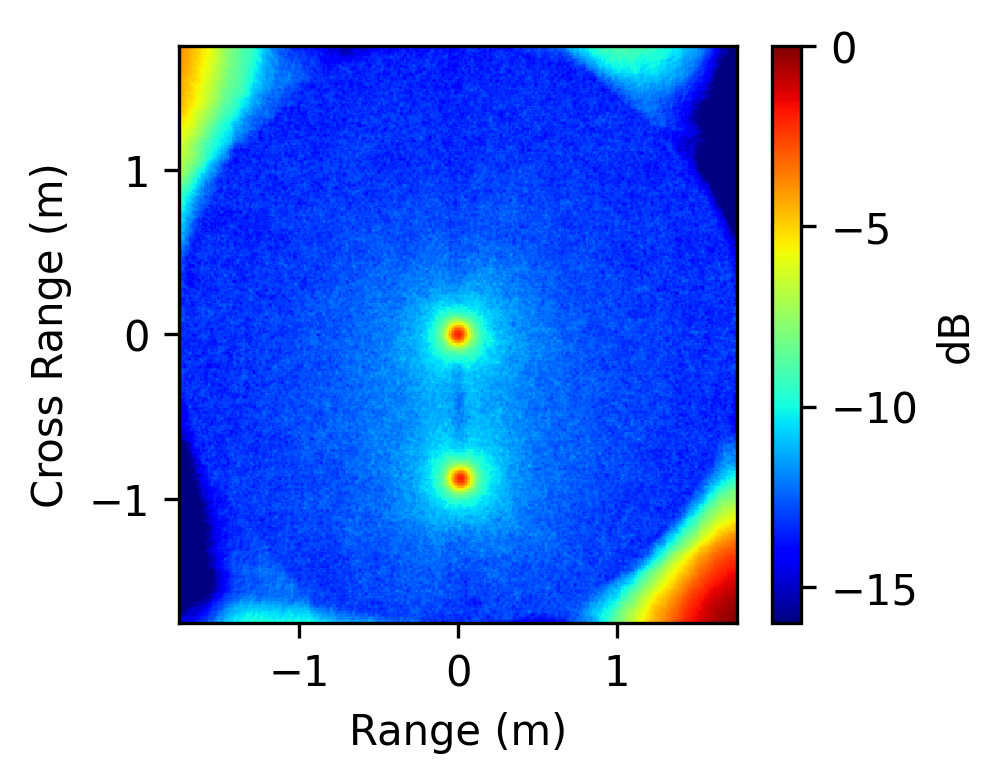} & 
        \includegraphics[scale=0.3]{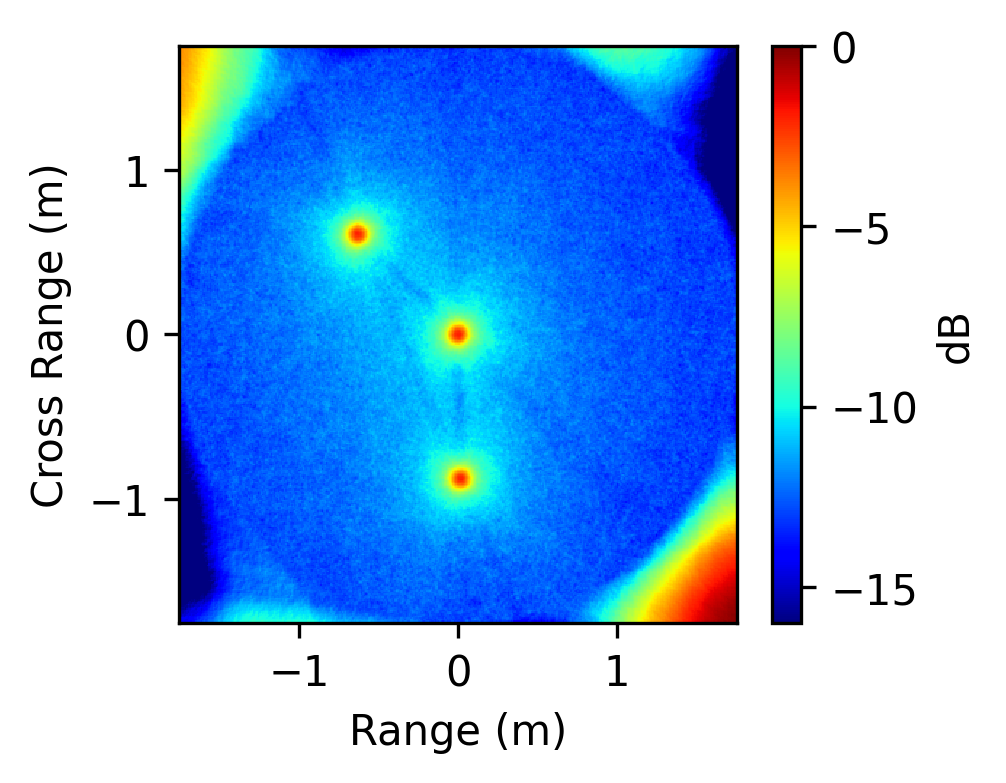} & 
        \includegraphics[scale=0.3]{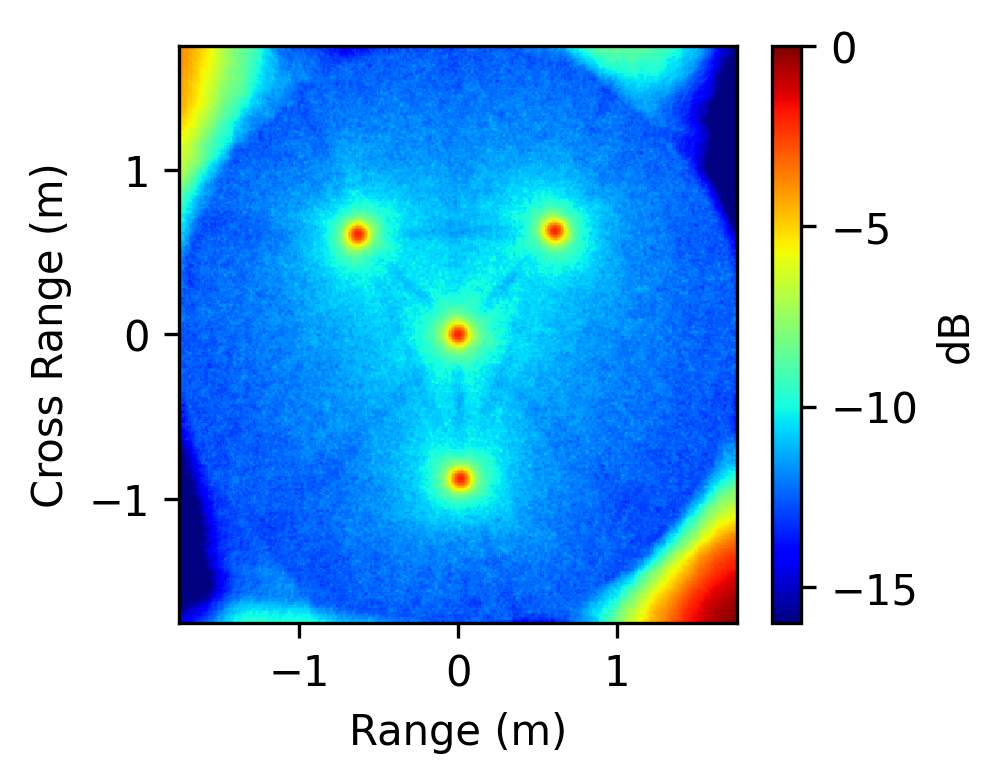} \\
        
        \parbox[c]{\hsize}{\rotatebox{90}{\centering \scriptsize \textbf{ATS}}} &
        \includegraphics[scale=0.3]{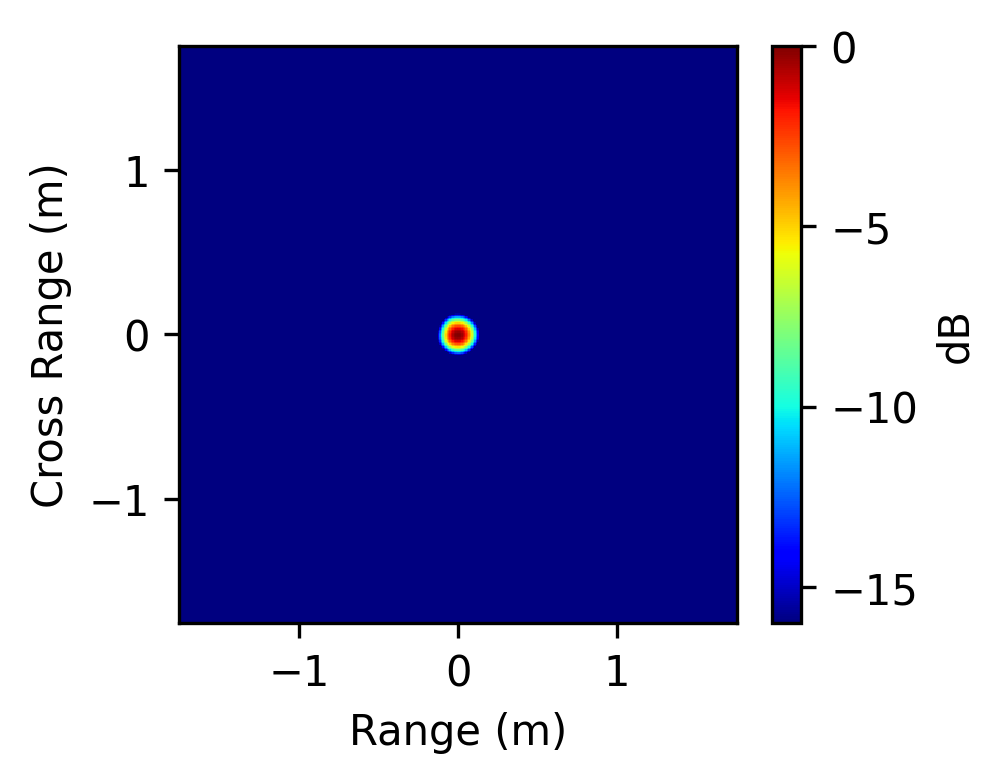} & 
        \includegraphics[scale=0.3]{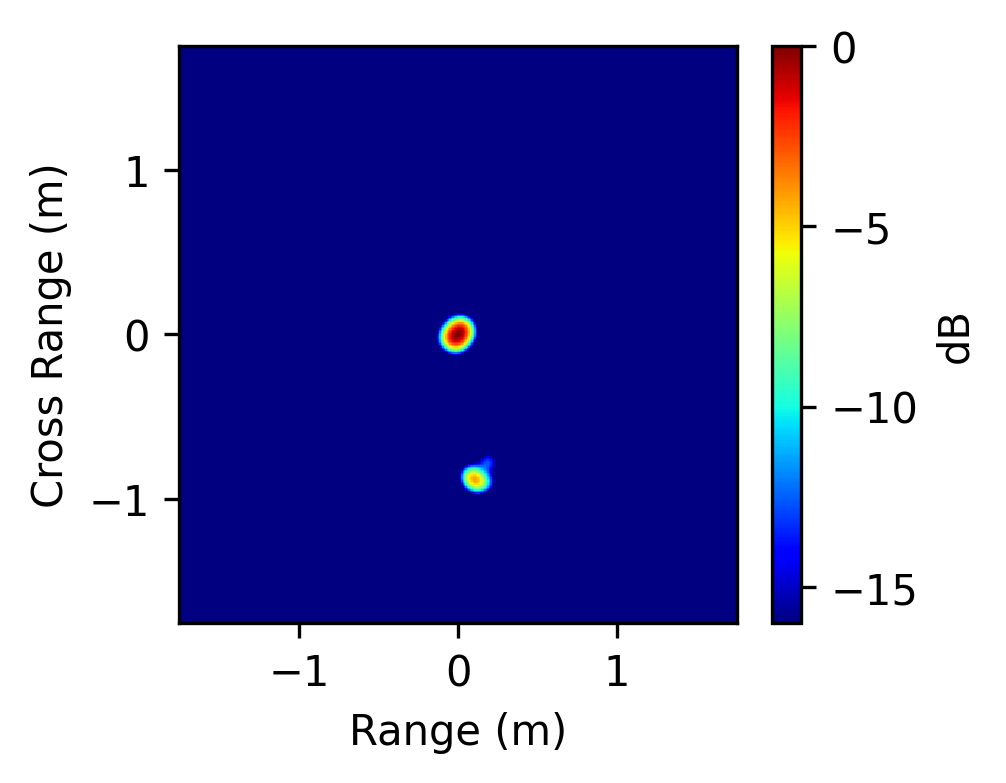} & 
        \includegraphics[scale=0.3]{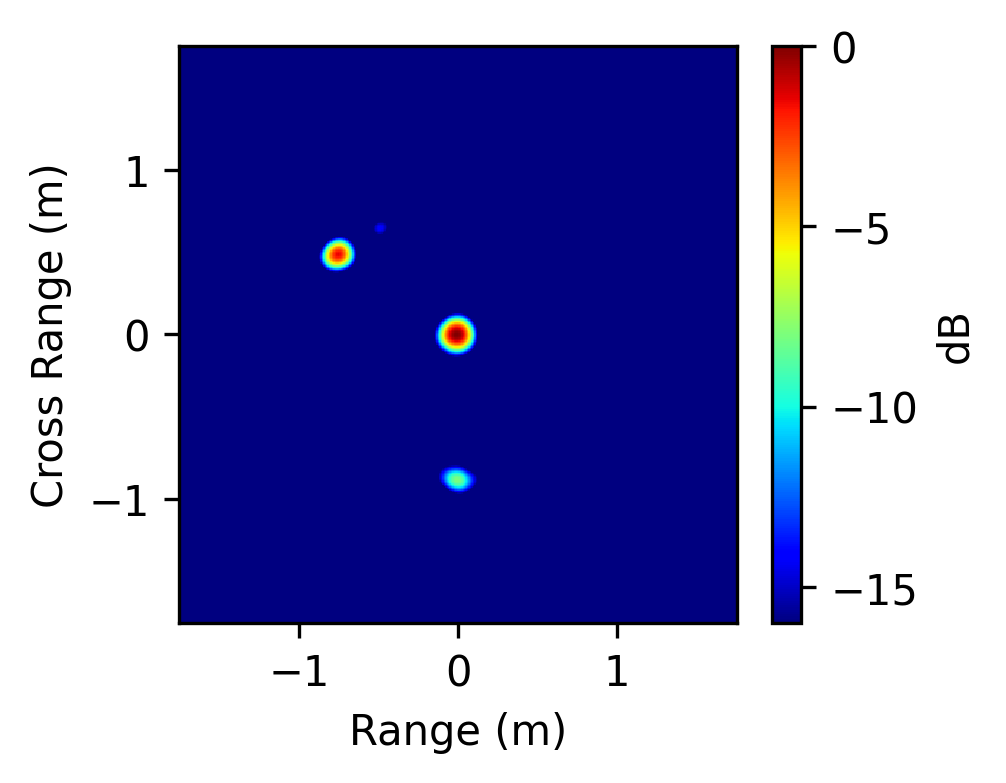} & 
        \includegraphics[scale=0.3]{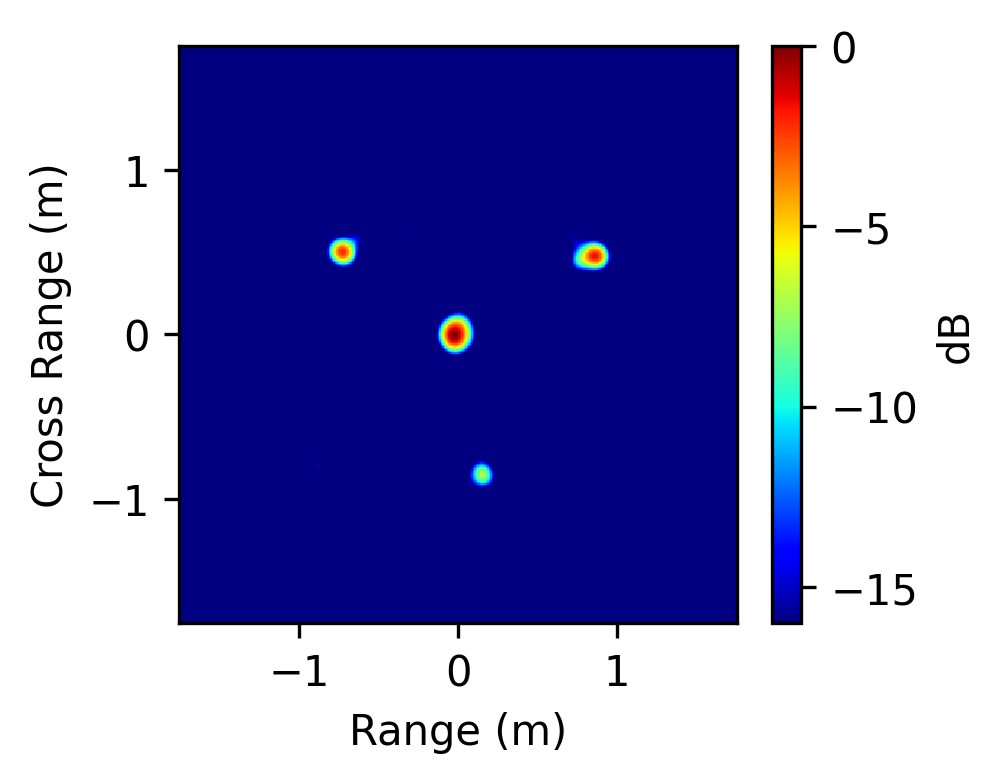} \\
    \end{tabular}
\end{minipage}

\caption{Simulated data in noise-free and noisy conditions. The rows show the Sinogram, BP, and ATS reconstructions respectively for single, double, triple, and quad point targets.}
\label{fig: simulated_data}
\end{figure*}

\section{Synthetic Data Experiments}

\begin{table} [htb!]
  \caption{Quantitative Metrics Comparison between different Reconstruction Methods on Simulated Data with added Gaussian Noise $\text{CN} \sim \mathcal{N}(0, 0.1)$}~\label{tab: quant_metrics_compare}
  \label{tab: quant_comparison}
  \centering
  \begin{tabular}{  c | l |c| c | c }
         \# Reflectors & \textit{Method} &\textit{PSNR $\uparrow$} &  \textit{LPIPS $\downarrow$} & \textit{MSE $\downarrow$ }  \\ 
         \hline
         & \textit{BP}              & 11.797  & 1.055  & 0.062 \\
         One & \textit{ATS}  & \textbf{15.049}  & \textbf{0.069}  & \textbf{0.001} \\ 
         \hline
         & \textit{BP}              & 11.296  & 1.050  & 0.069\\
         Two & \textit{ATS}  & \textbf{14.611}  & \textbf{0.047}  & \textbf{0.004} \\ 
         \hline
         & \textit{BP}              & 10.754  & 1.043  & 0.077 \\
         Three & \textit{ATS}  & \textbf{14.134}  & \textbf{0.069}  & \textbf{0.004} \\ 
         \hline
         & \textit{BP}              & 10.266  & 1.035  & 0.084 \\
         Four & \textit{ATS}  & \textbf{13.746}  & \textbf{0.118}  & \textbf{0.004} \\ 
         
         \hline
        
  \end{tabular}
\end{table}
To generate synthetic radar sinograms depicting various point targets, we employ Equation \ref{eqn: radar_signal}. Four distinct synthetic scenarios were simulated, each containing single, double, triple, and quadruple targets within the scene. The corresponding results are presented in Figure \ref{fig: simulated_data}. ATS demonstrates qualitative consistency with BP, exhibiting accurate scene reconstruction in noise-free conditions and superior performance in noisy conditions. BP, on the other hand, shows strong artifacts around the corners of each reconstructed scene when noise is added.

\subsection{Effect of Noise}

To quantitatively evaluate our proposed ATS reconstruction against BP, we use three image metrics namely- PSNR (Peak Signal to Noise Ratio), LPIPS (Learned Perceptual Image Patch Similarity) \cite{zhang2018unreasonable}, and MSE (Mean Squared Error). Table \ref{tab: quant_comparison} presents the comparison of quantitative metrics among various reconstruction methods on four different simulated scenes with added Gaussian noise $\text{CN} \sim \mathcal{N}(0, 0.1)$ per scan. As shown in Table \ref{tab: quant_comparison}, the proposed ATS consistently outperforms BP across all metrics in noisy scenes.

\subsection{Effect of Sparse Measurements}

\begin{figure}
  \begin{subfigure}[b]{1\columnwidth}
    \includegraphics[width=\linewidth]{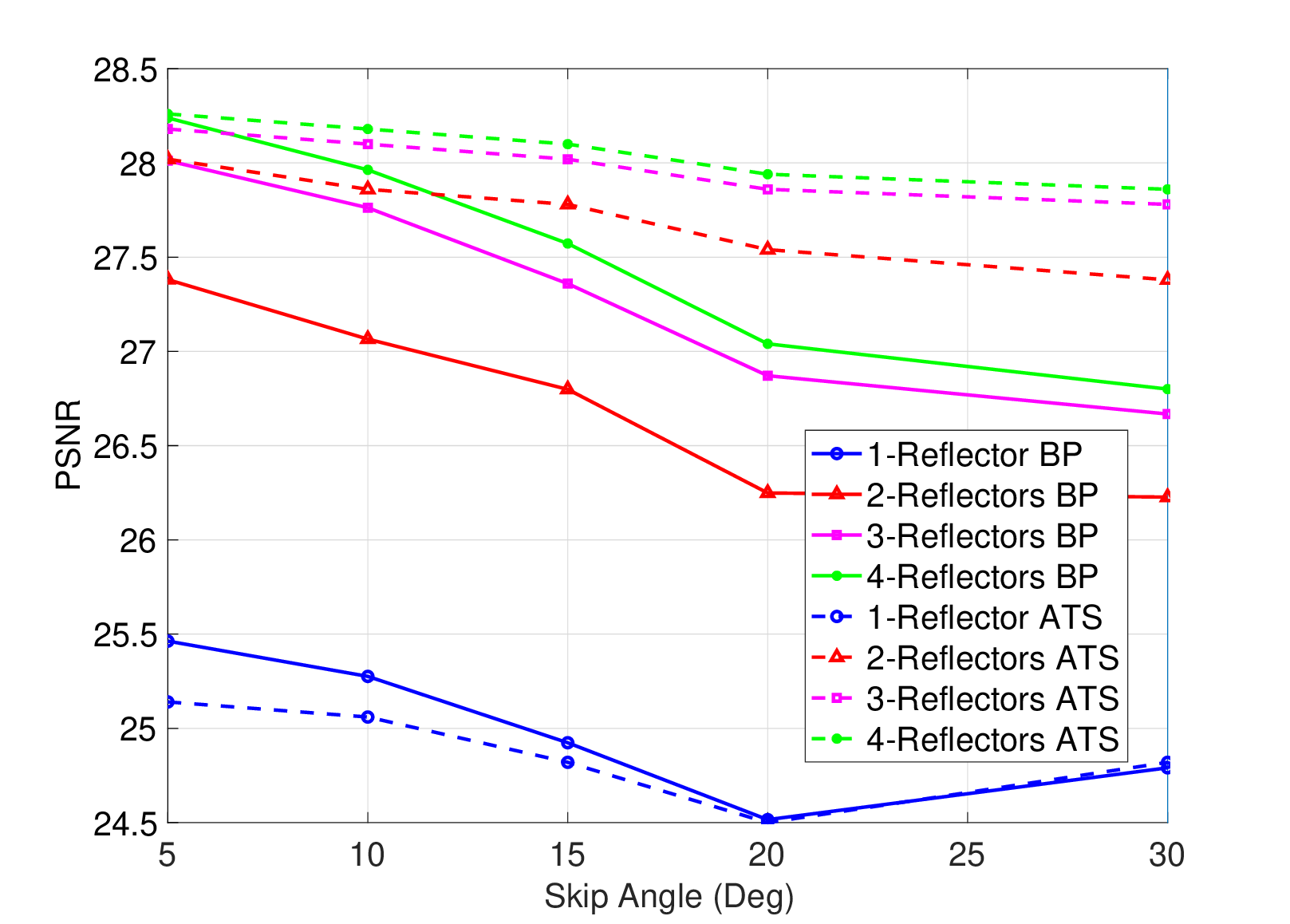}
    \caption{PSNR $\uparrow$ comparison for BP and ATS}
    \label{fig: psnr}
  \end{subfigure}
  
  \begin{subfigure}[b]{1\columnwidth}
    \includegraphics[width=\linewidth]{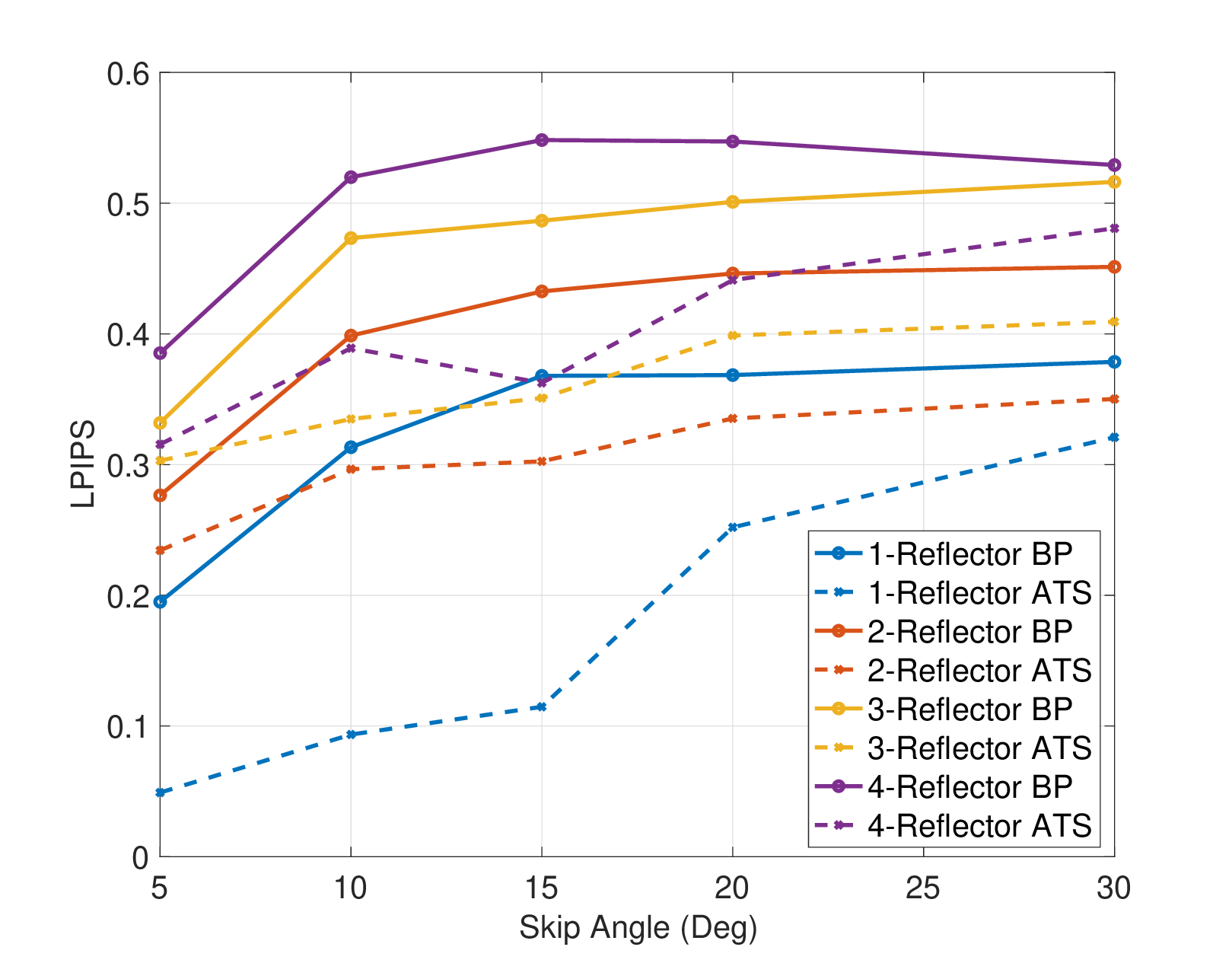}
    \caption{LPIPS $\downarrow$ comparison for BP and ATS}
    \label{fig: lpips}
  \end{subfigure}
  \caption{Comparison of PSNR and LPIPS metrics for BP and ATS with different numbers of simulated reflectors at various skip angles}
\end{figure}

Figures \ref{fig: psnr} and \ref{fig: lpips} illustrate the influence of sparse measurements in terms of skip angles on PSNR and LPIPS metrics, respectively, for both BP and ATS across varying numbers of simulated targets. With an increasing number of targets, PSNR tends to decrease for both BP and ATS; however, ATS consistently maintains higher PSNR values compared to BP across all skip angles except for the single reflector case where the PSNR difference is less than 0.4. Similarly, ATS exhibits lower LPIPS values compared to BP across all skip angles, indicating superior performance in scenarios with limited measurements.

\section{Real Data Experiments}

\begin{figure}
\begin{minipage}{\textwidth}
    \begin{tabular} {c|c|c}
        \scriptsize \textbf{Target} & \scriptsize \textbf{BP} & \scriptsize \textbf{ATS} \\
        \hline
        \includegraphics[scale=0.017]{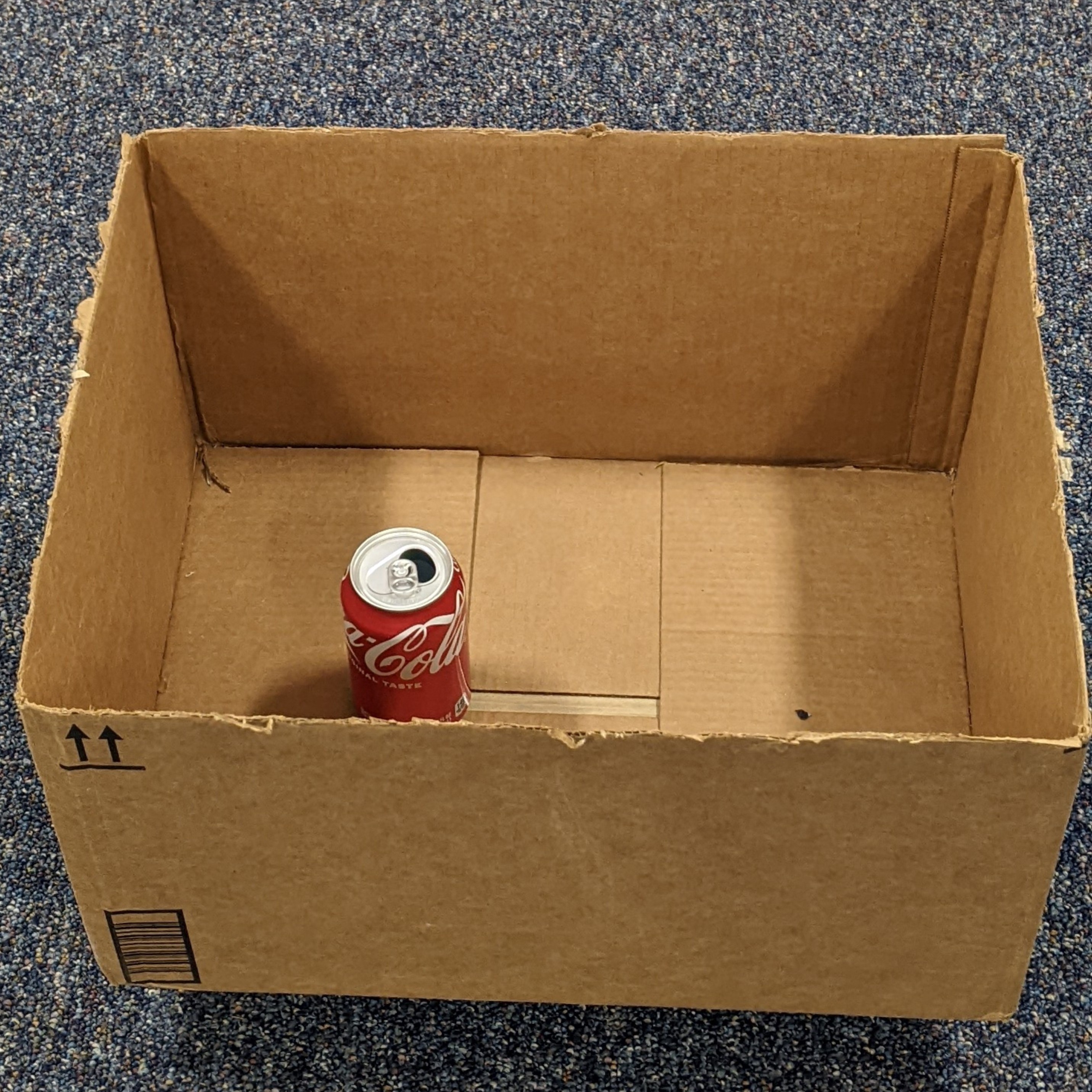}  & 
        \includegraphics[scale=0.3]{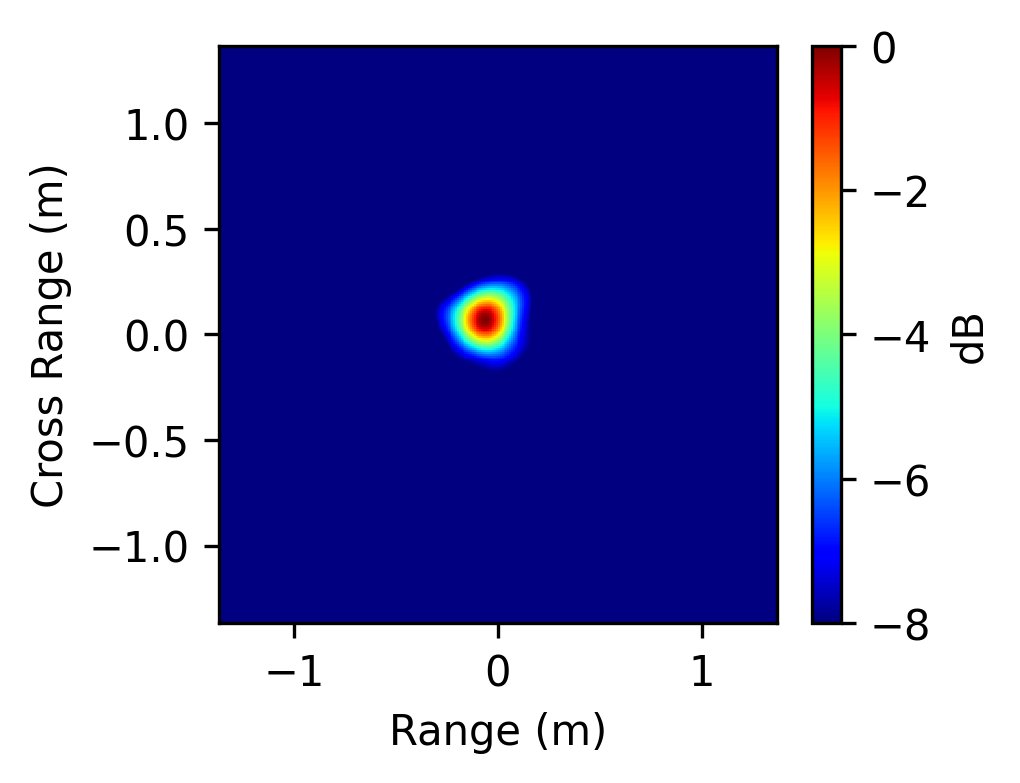} & 
        \includegraphics[scale=0.3]{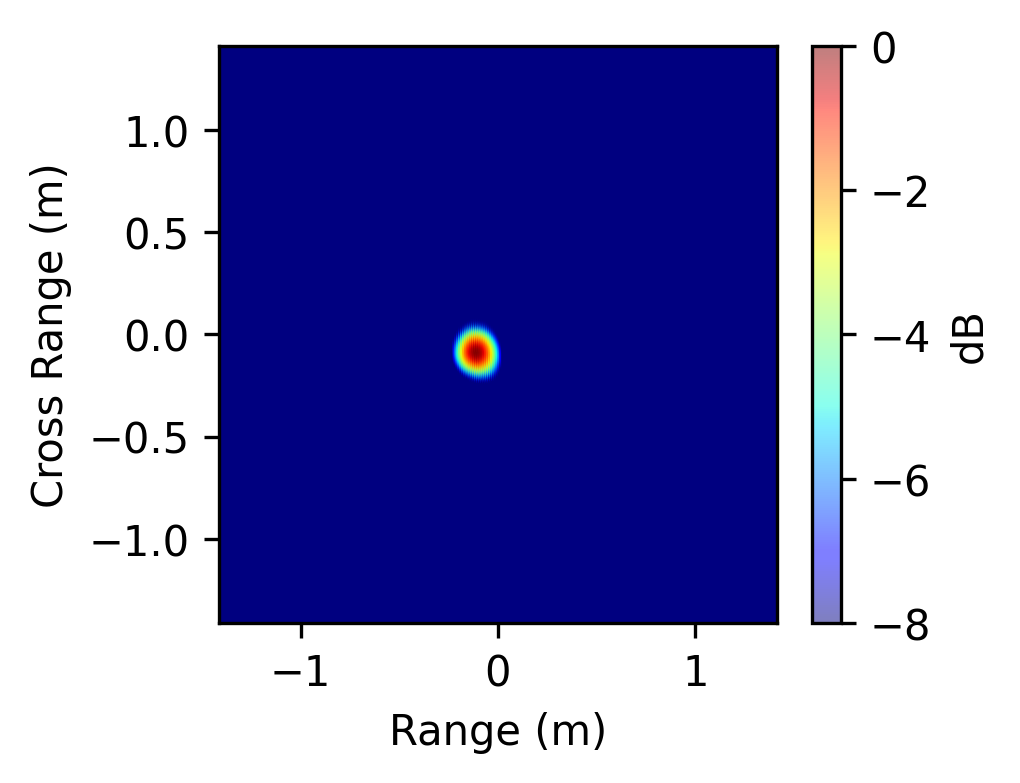} \\
        \hline

        \includegraphics[scale=0.0165]{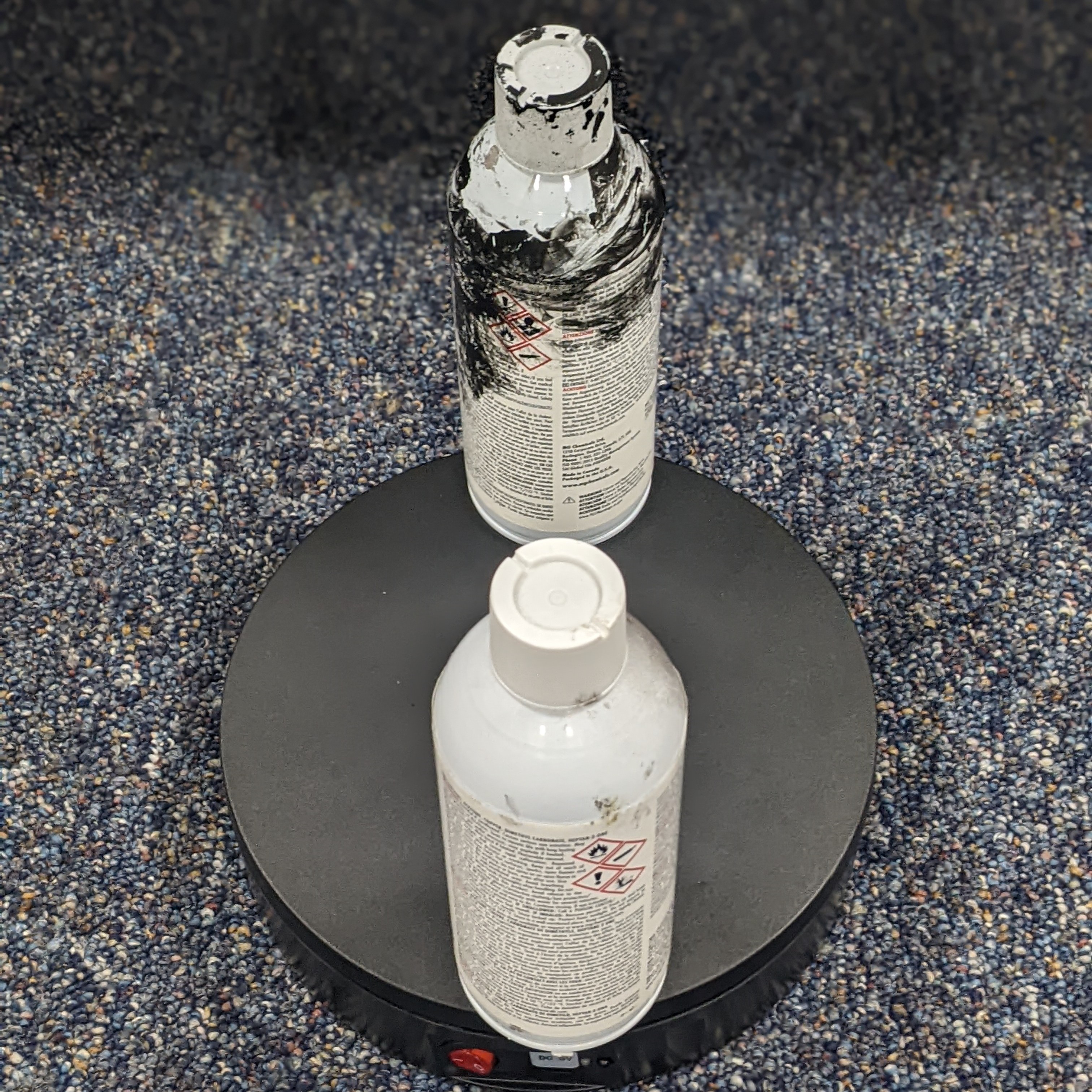} & 
        \includegraphics[scale=0.3]{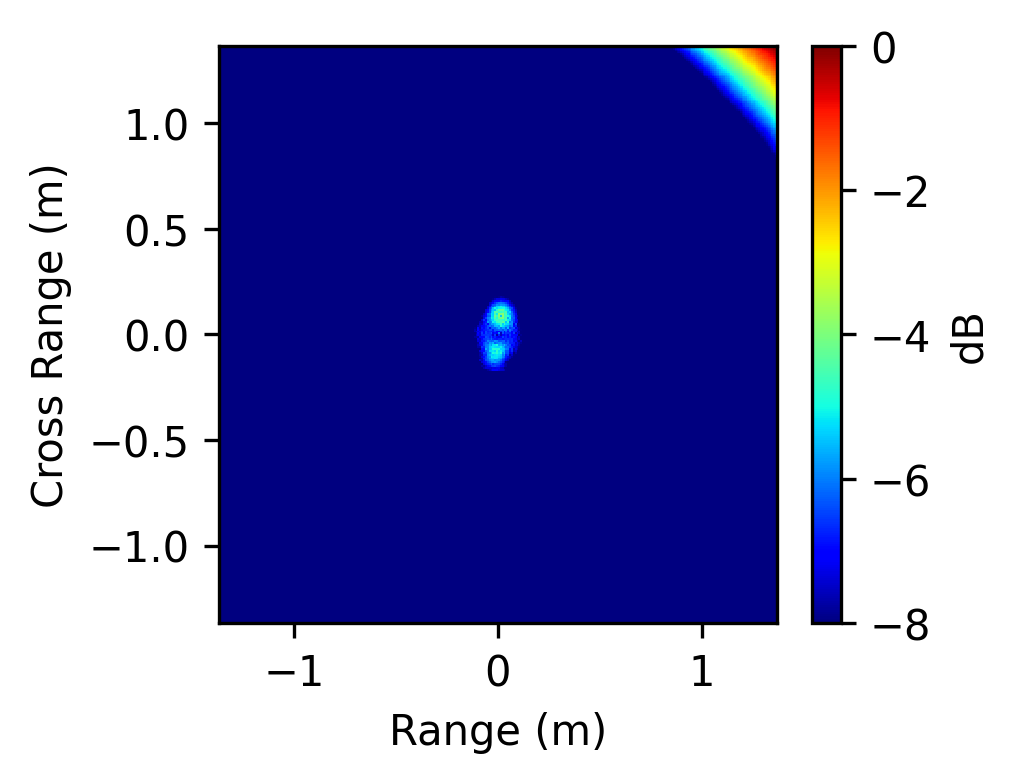} & 
        \includegraphics[scale=0.3]{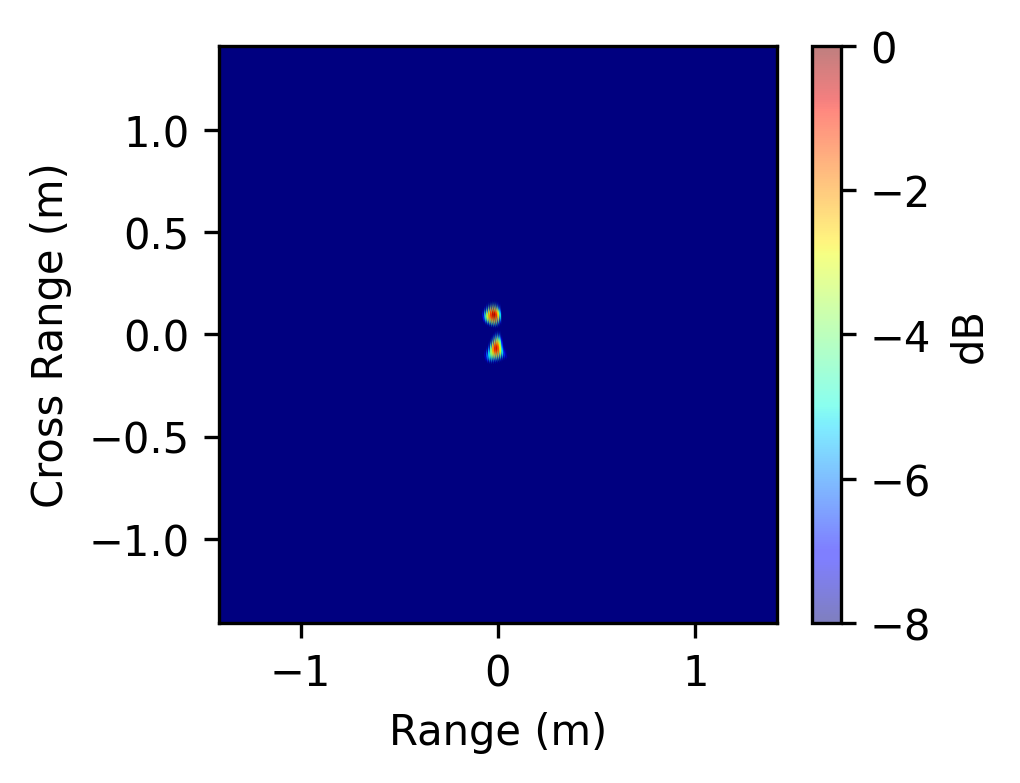} \\
        \hline
        \includegraphics[scale=0.0165]{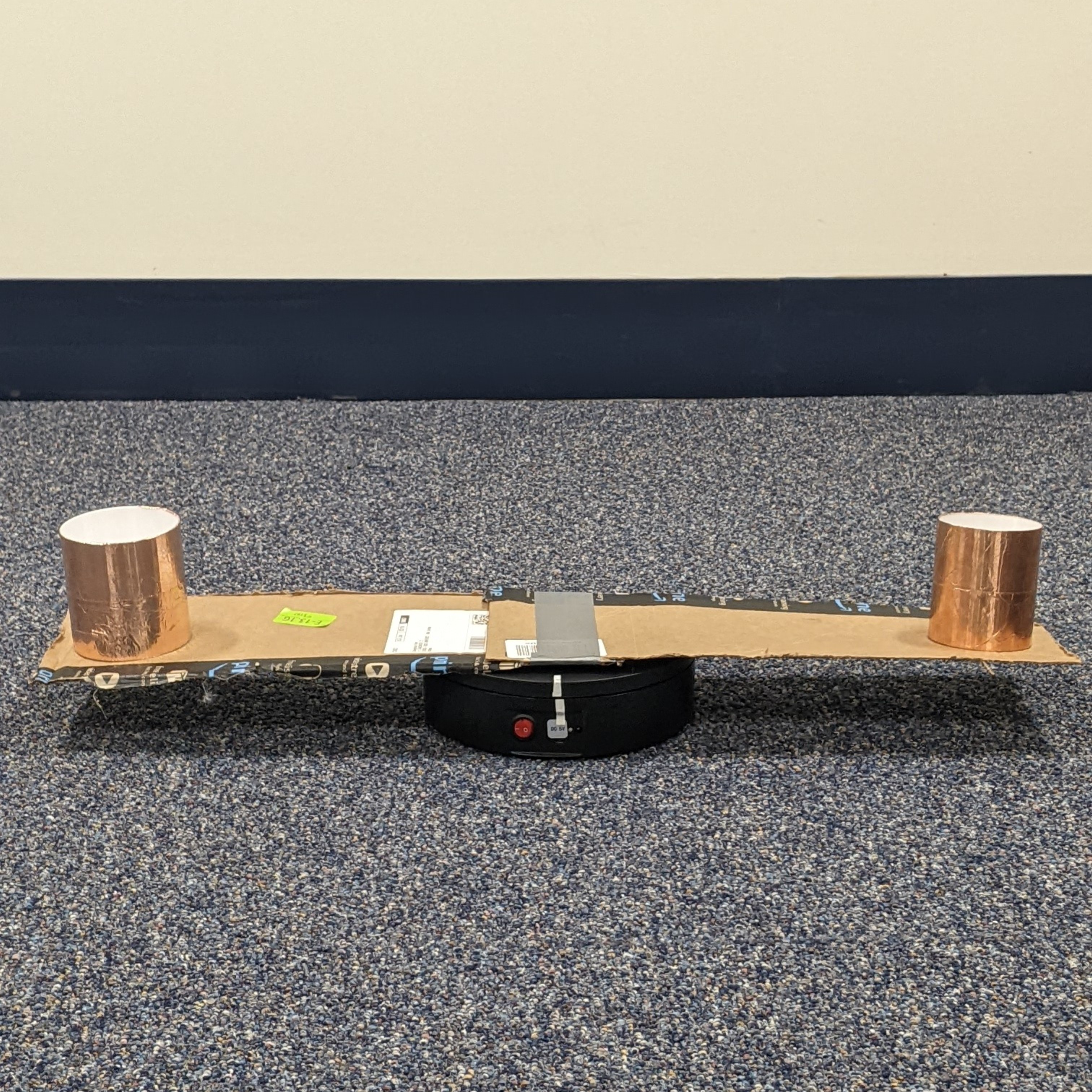} & 
        \includegraphics[scale=0.3]{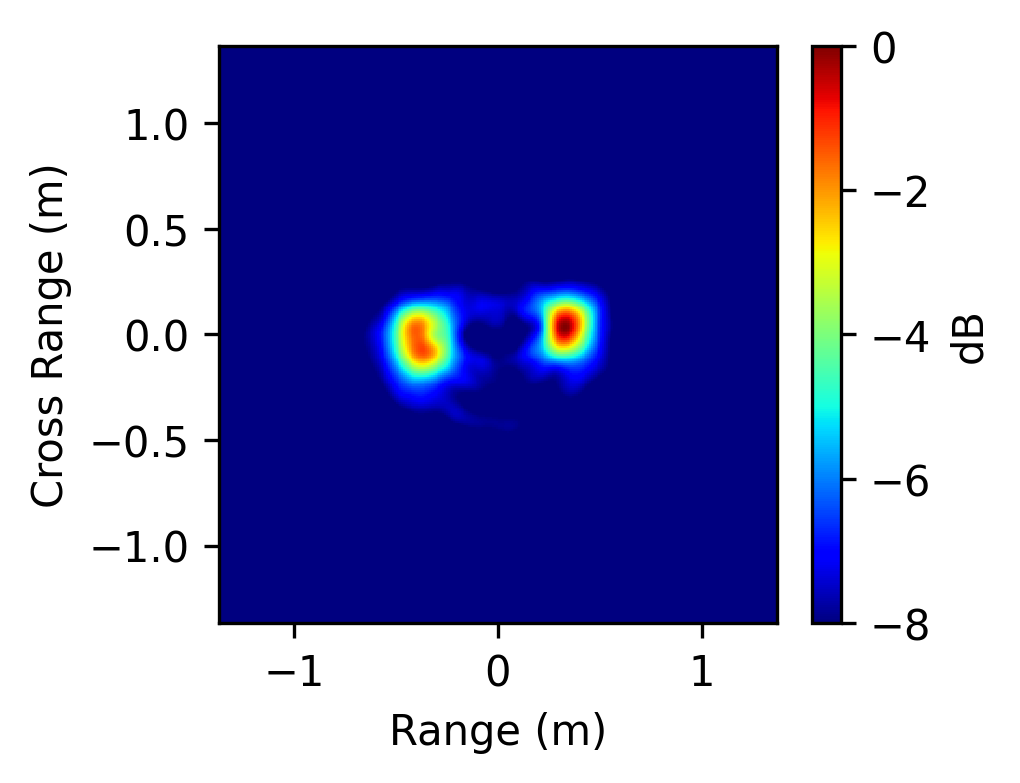} & 
        \includegraphics[scale=0.3]{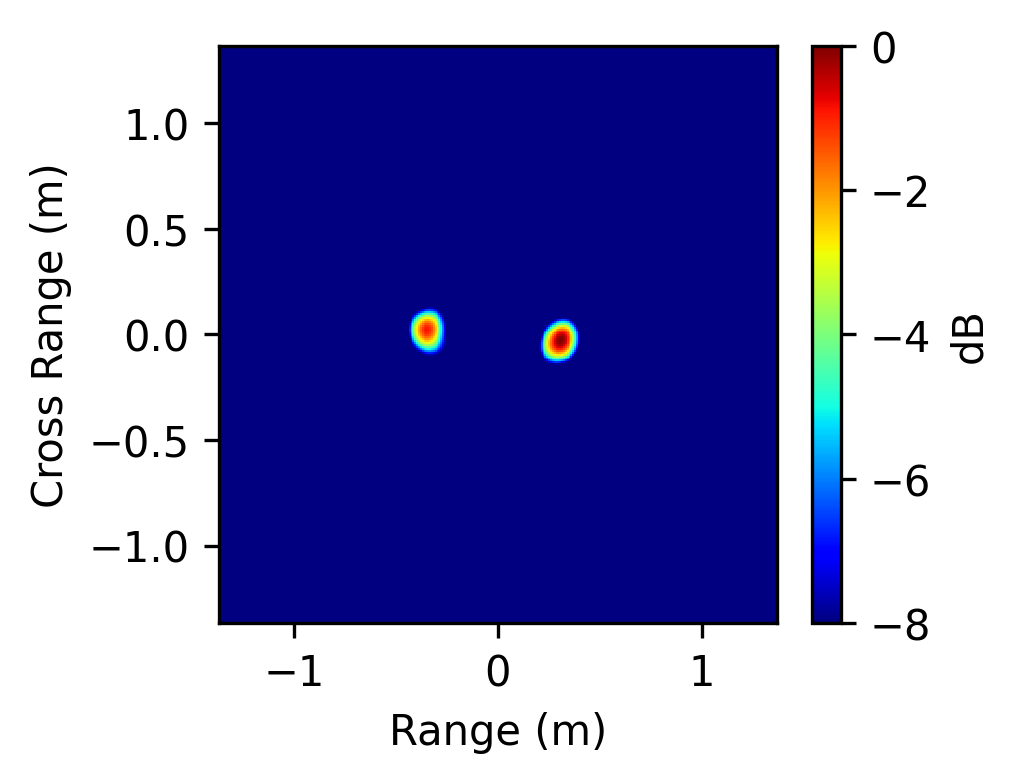} \\
        \hline
        \includegraphics[scale=0.021]{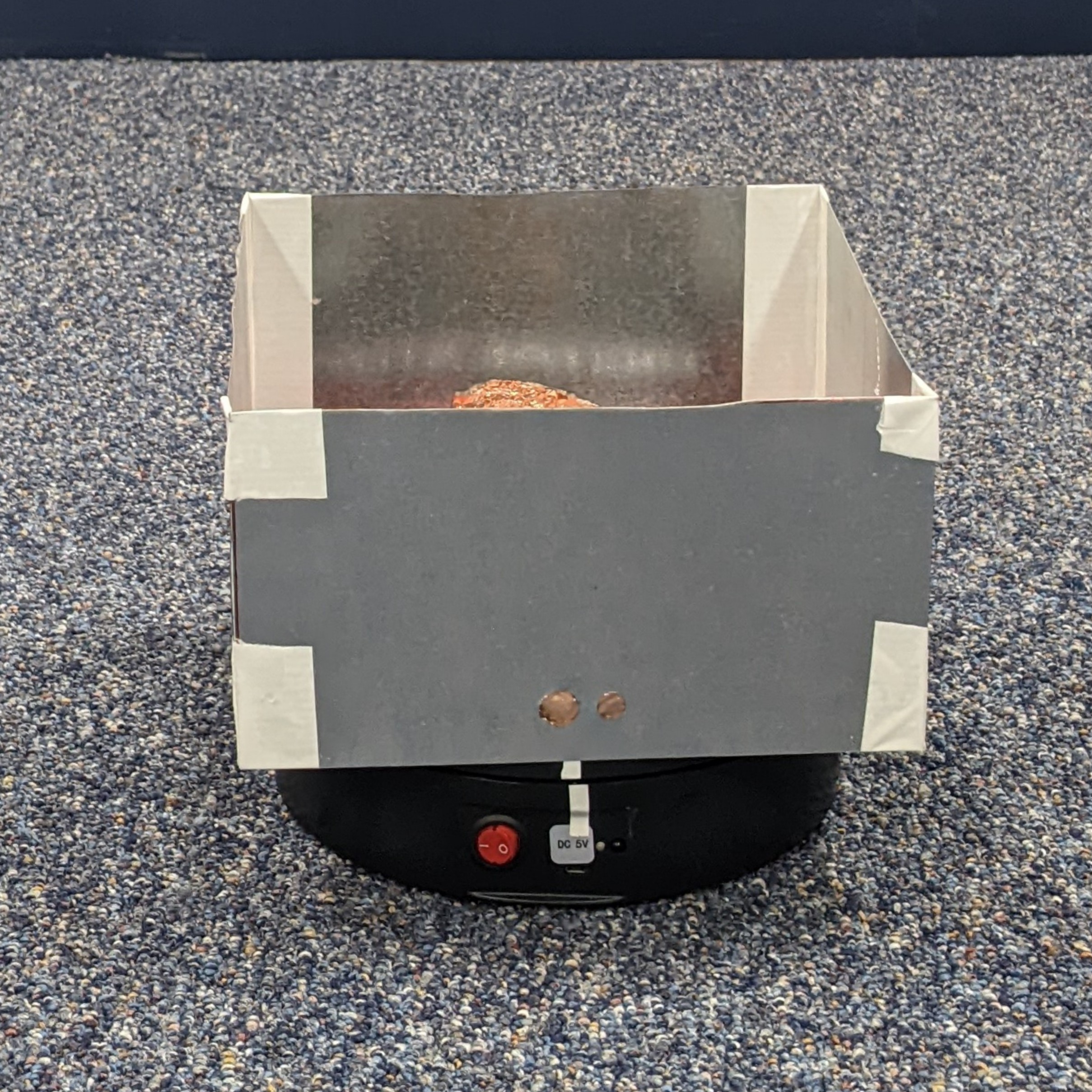} & 
        \includegraphics[scale=0.3]{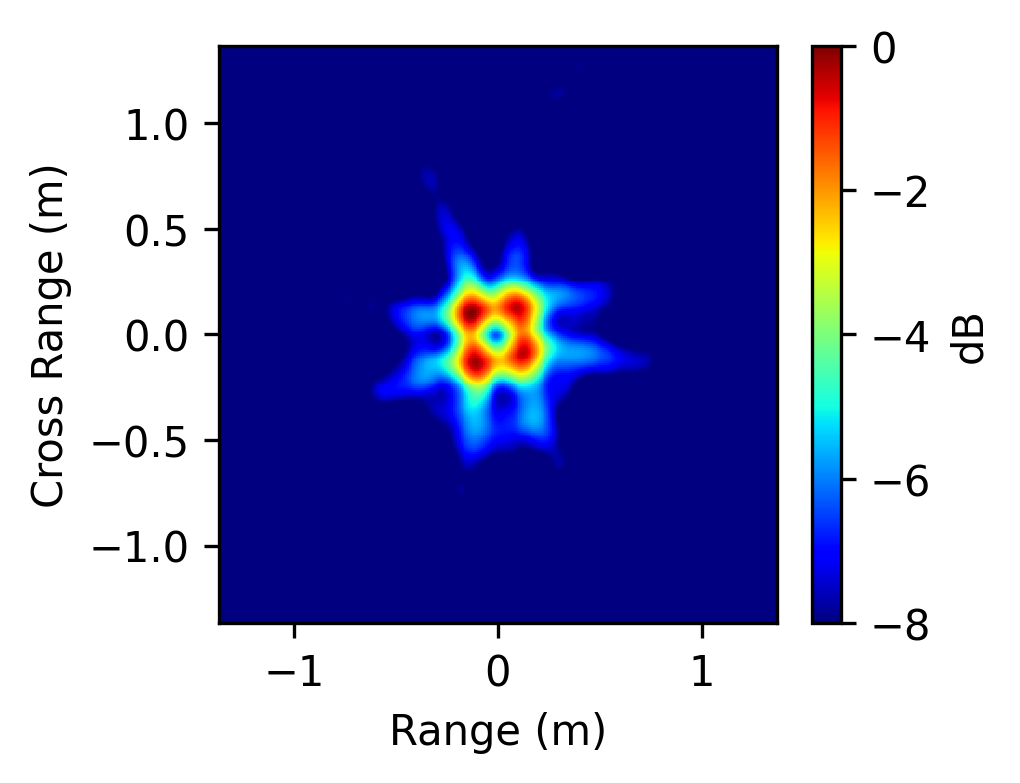} & 
        \includegraphics[scale=0.3]{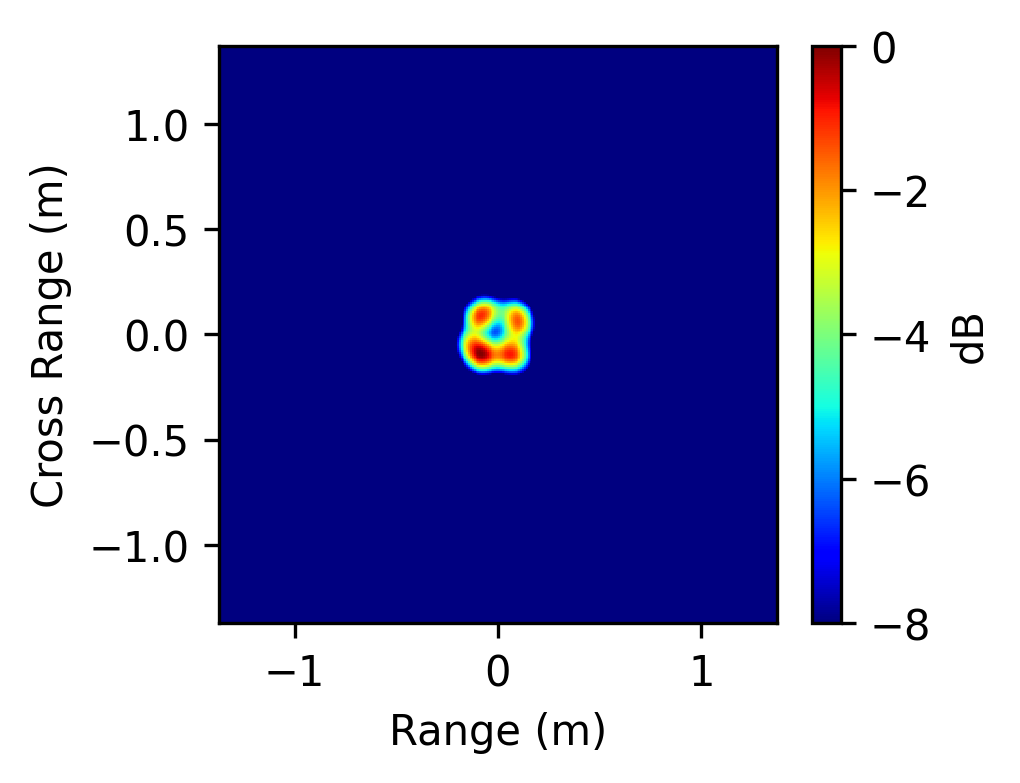} \\
        \hline
    \end{tabular}
\end{minipage}

\caption{Real data (with full $360\degree$ views with 1 scan per viewing angle) - Row 1: single soda can inside a cardboard box, Row 2: two spray paint bottles 7 inches (17.78 cm) apart, Row 3: double cans 27 inches (68.58 cm) apart, Row 4: square metal box with a side length of 9.5 inches (24.13 cm).}
\label{fig: real_data}
\end{figure}

Figure \ref{fig: real_data}  illustrates ISAR imaging results obtained with real data collected with - a single cylindrical object, two cylindrical objects placed at varying distances apart, and a square container. The second and third columns of the figure illustrate reconstructed images using BP and ATS, respectively. BP begins to exhibit artifacts as the complexity of the target scene increases. In contrast, ATS reconstruction displays no artifacts and maintains the reconstructed targets' dimensions more accurately to their original form.

\subsection{Effect of Skip Angles on Real Scene Reconstruction}
The impact of measurements with skip angles on the reconstruction algorithms for imaging double soda cans is depicted in Figure \ref{fig: skip_angles}. Increasing skip angles introduces significant artifacts in BP reconstruction, whereas ATS reconstruction only suffers from minor amplitude attenuation.

\subsection{Effect of Partial Rotations on Real Scene Reconstruction}
The effect of partial rotation of the target on BP and ATS reconstructions of double soda cans is shown in Figure \ref{fig: partial_angles}. It can be observed that BP fails to reconstruct the secondary object when the measurements are less than or equal to half of the total 360 measurement angles.

\newcolumntype{M}[1]{>{\centering\arraybackslash}m{#1}}

\begin{figure}
\centering
\begin{tabular}{M{2cm}|M{2.5cm}|M{2.5cm}}
    \scriptsize \textbf{Skip Angles} & \scriptsize \textbf{BP} & \scriptsize \textbf{ATS} \\
    \hline
    \scriptsize $10^\circ$ & \includegraphics[scale=0.3]{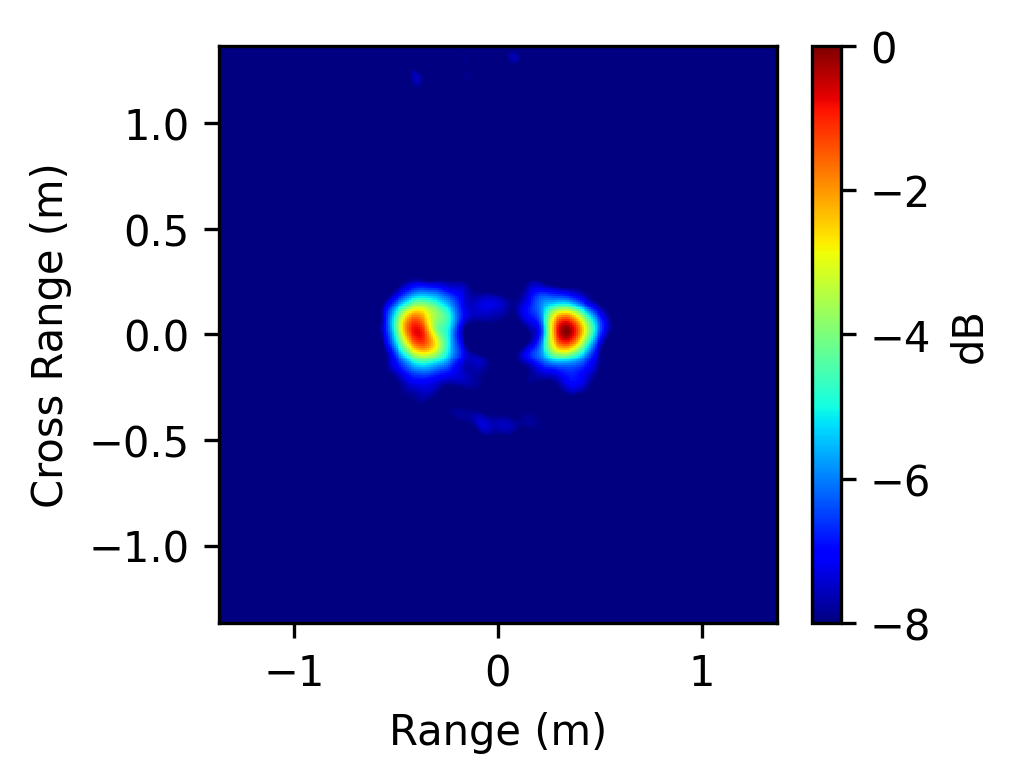} & \includegraphics[scale=0.3]{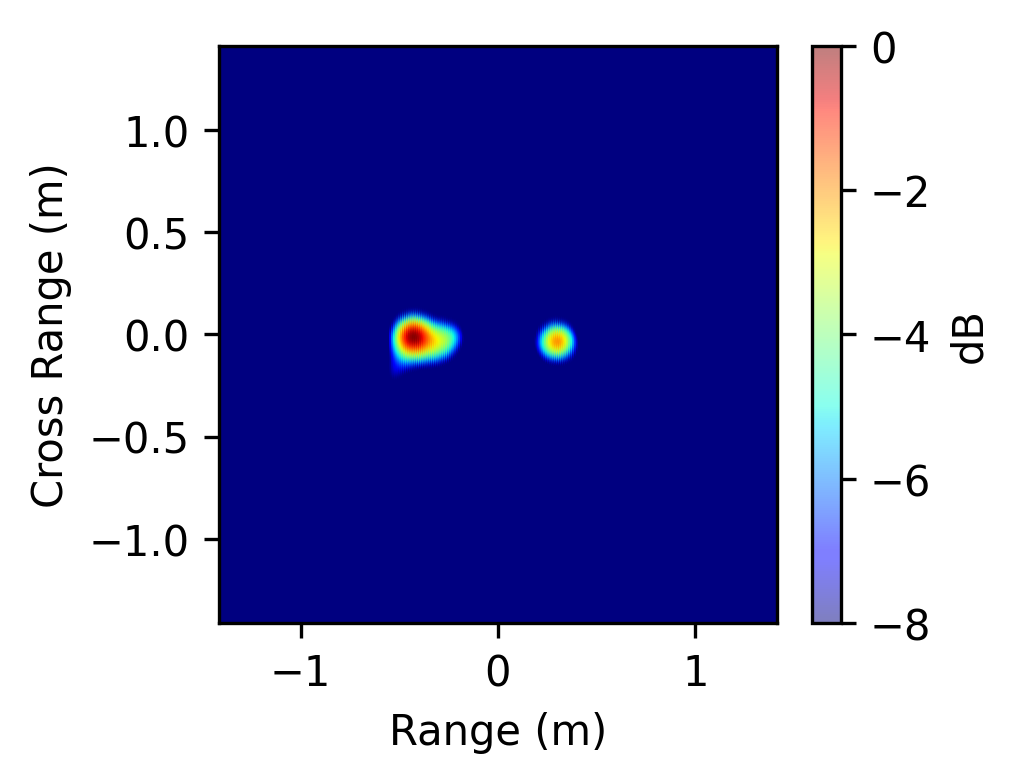} \\
    \hline
    \scriptsize $20^\circ$ & \includegraphics[scale=0.3]{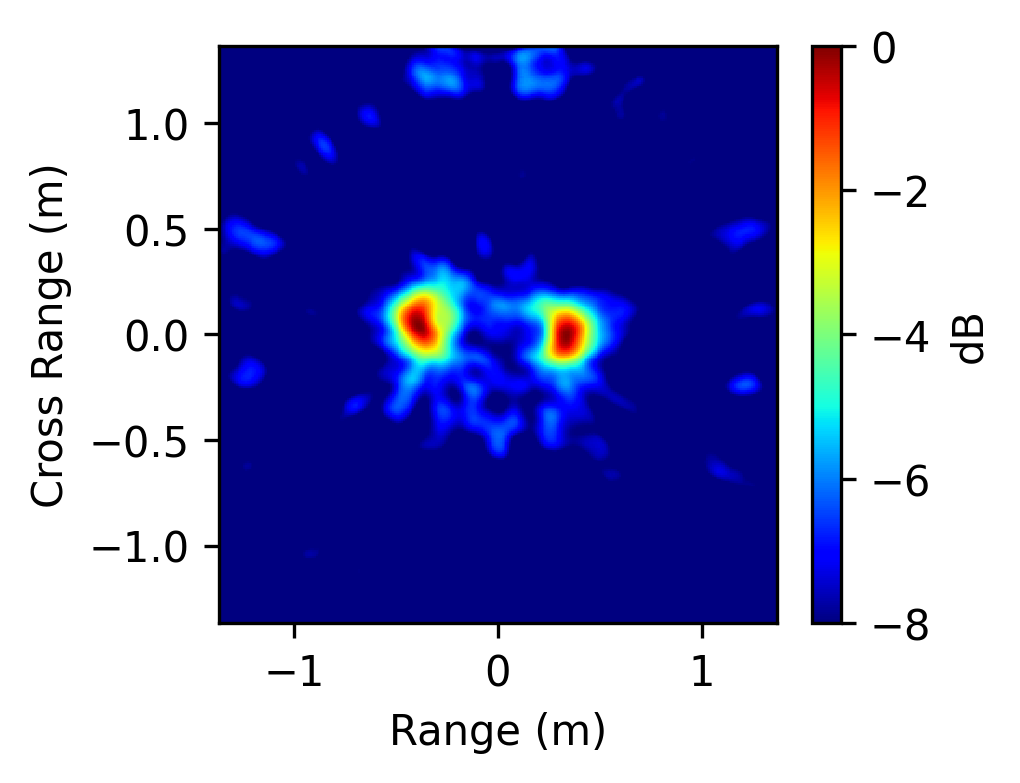} & \includegraphics[scale=0.3]{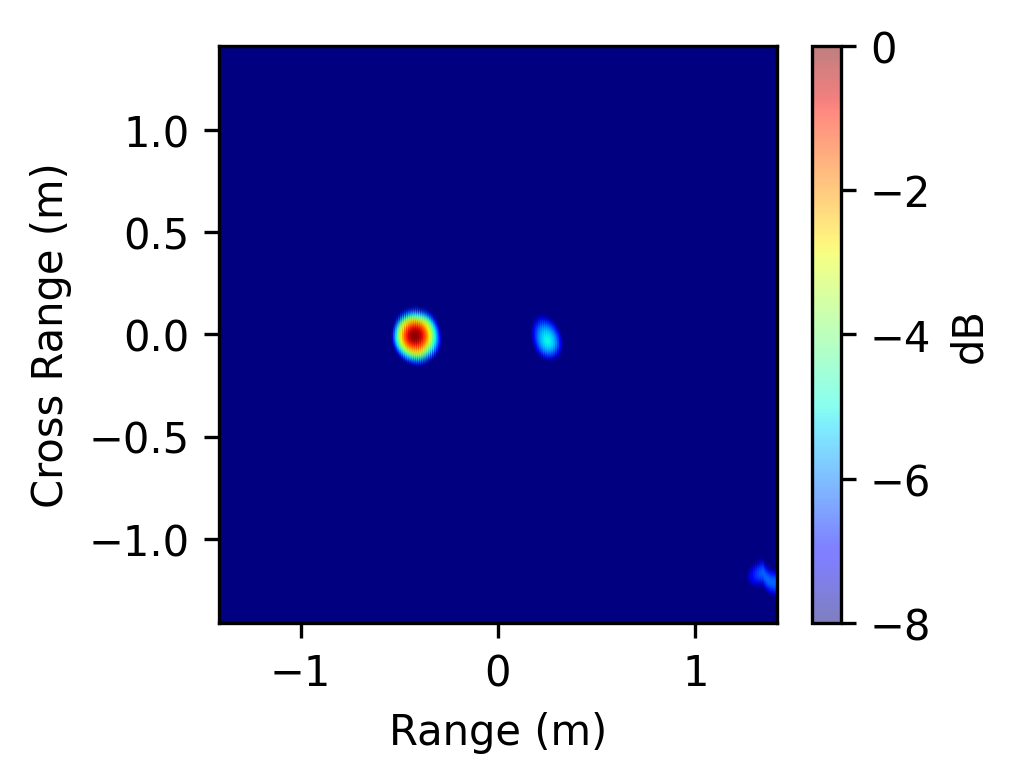} \\
    \hline
    \scriptsize $30^\circ$ & \includegraphics[scale=0.3]{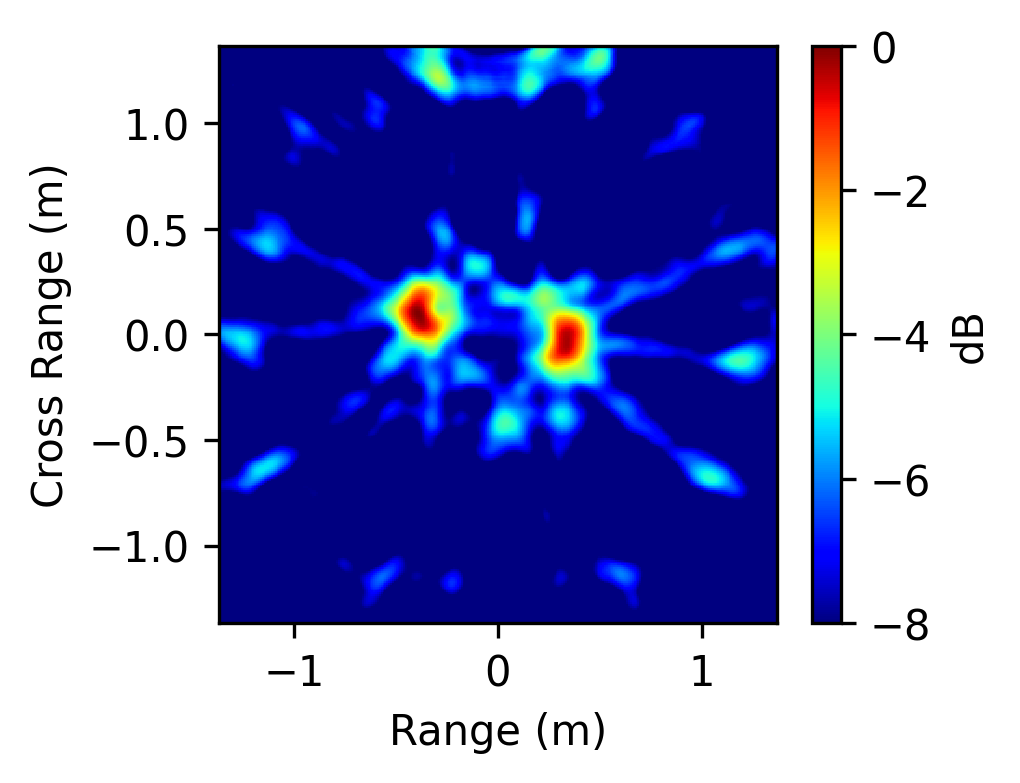} & \includegraphics[scale=0.3]{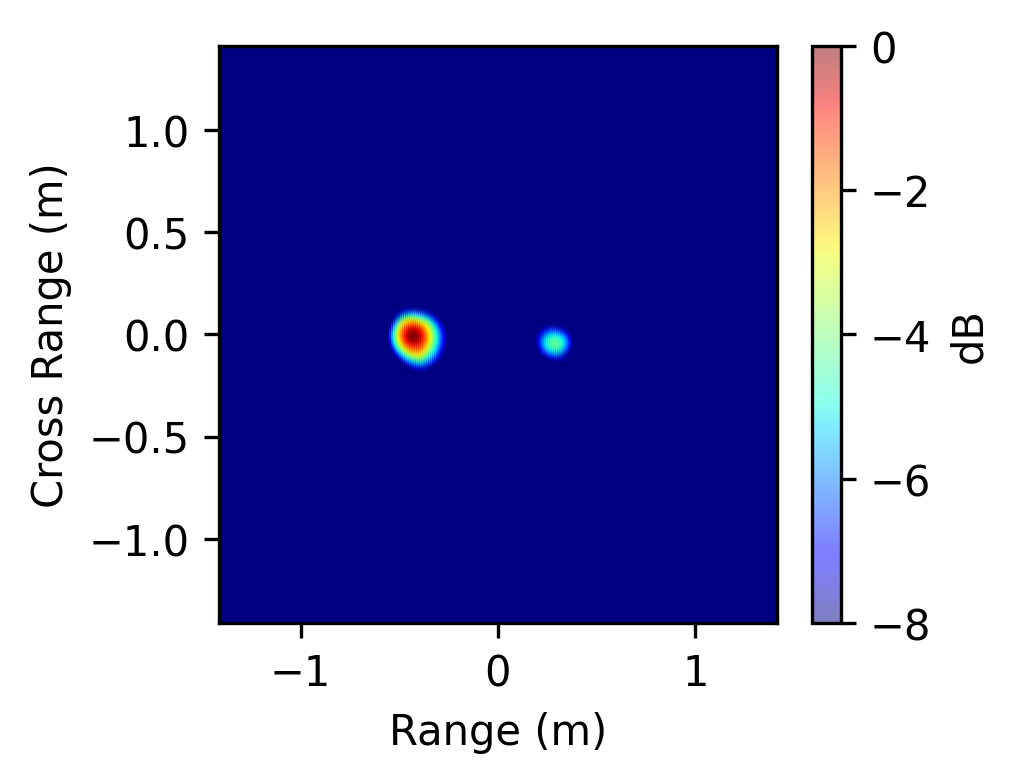} \\
    \hline
\end{tabular}

\caption{Impact of skip angles on double soda cans imaging. Row 1: $10^\circ$ skip (36 scans), Row 2: $20^\circ$ skip (18 scans), Row 3: $30^\circ$ skip (12 scans). Increasing skip angles leads to notable artifacts in BP. In contrast, ATS reconstruction experiences only a slight decrease in signal strength.}
\label{fig: skip_angles}
\end{figure}

\begin{figure}
\centering
\begin{tabular} [t!]{M{2.5cm}|M{2.4cm}|M{2.4cm}}
    \scriptsize \textbf{Partial Rotation} & \scriptsize \textbf{BP} & \scriptsize \textbf{ATS } \\
    \hline
     \scriptsize $0\degree$ to $90\degree$ & \includegraphics[scale = 0.3]{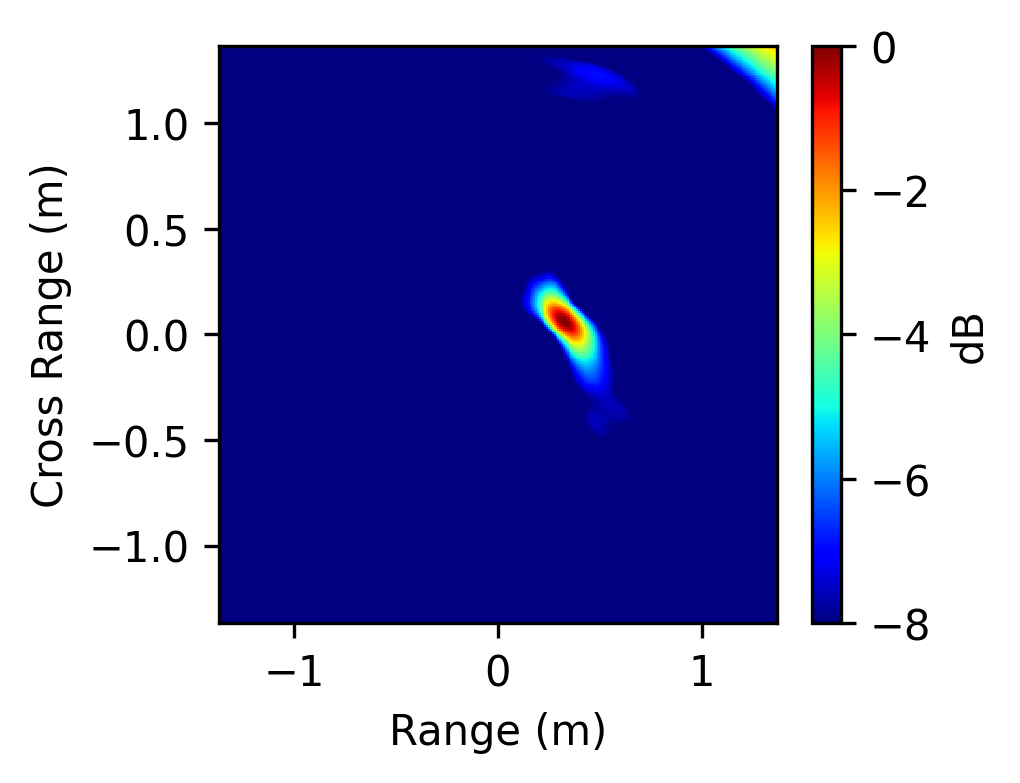} & \includegraphics[scale = 0.3]{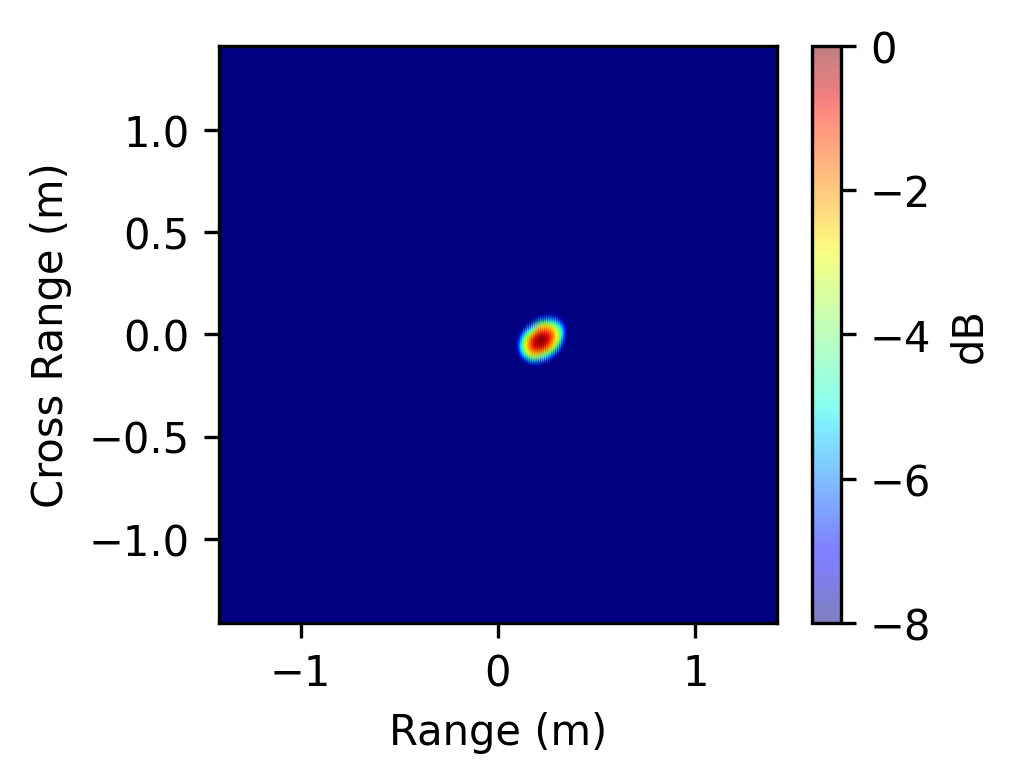} \\
    \hline
    \scriptsize $0\degree$ to $180\degree$ & \includegraphics[scale = 0.3]{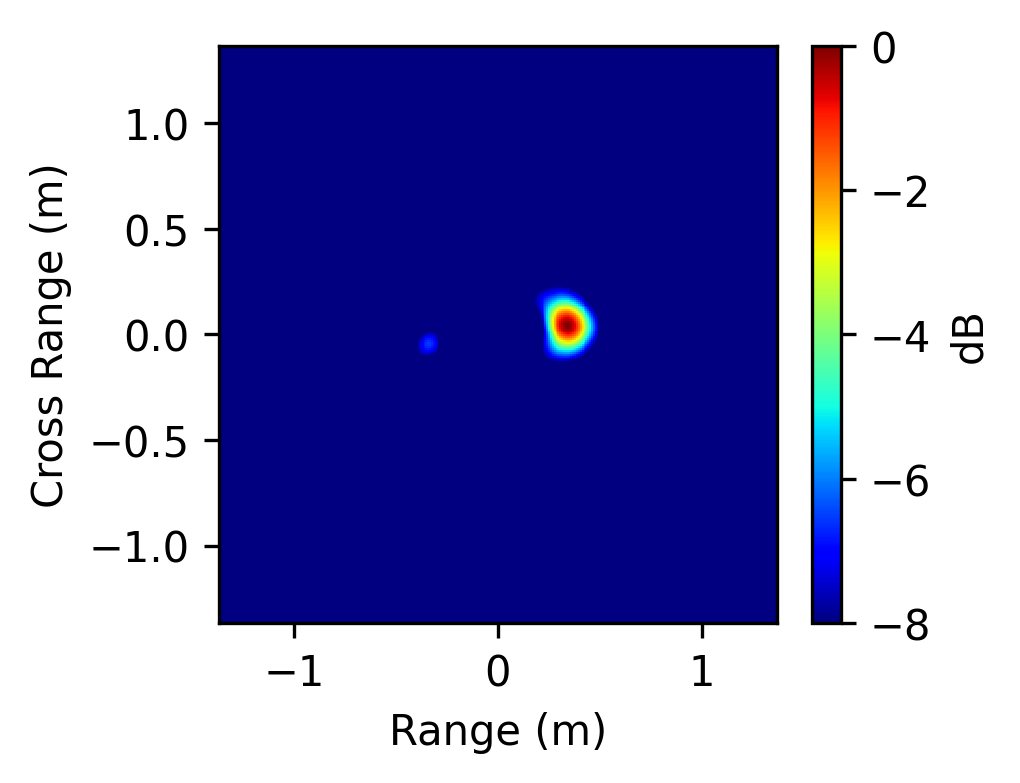} & \includegraphics[scale = 0.3]{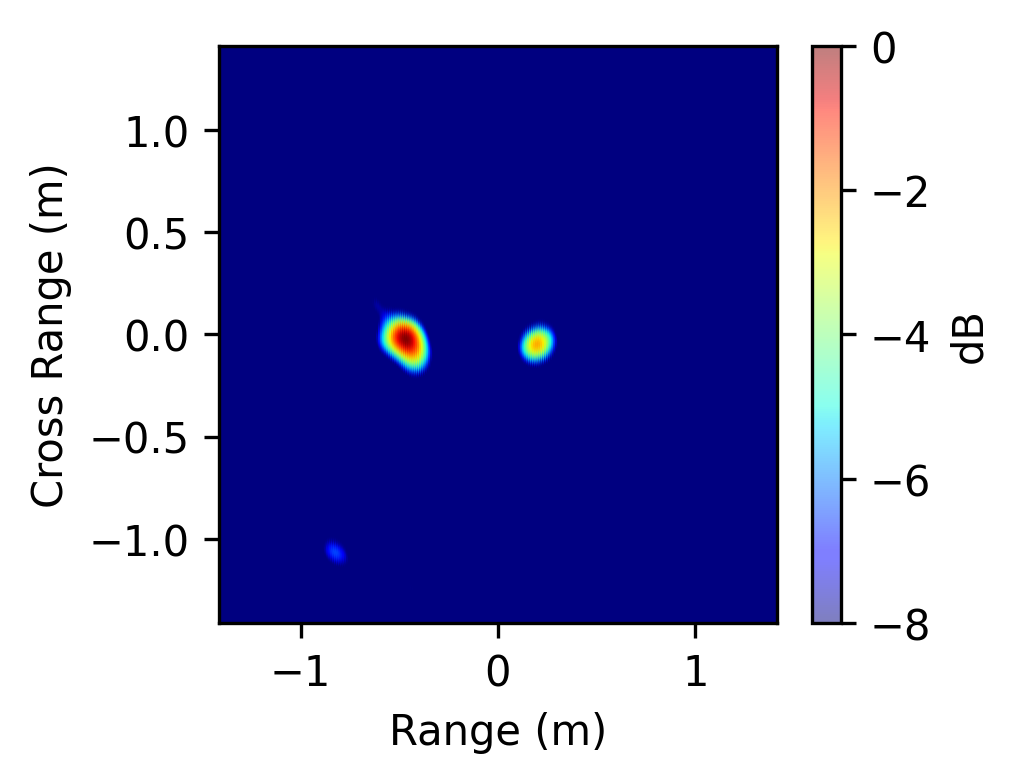} \\
    \hline
    \scriptsize $0\degree$ to $270\degree$ & \includegraphics[scale = 0.3]{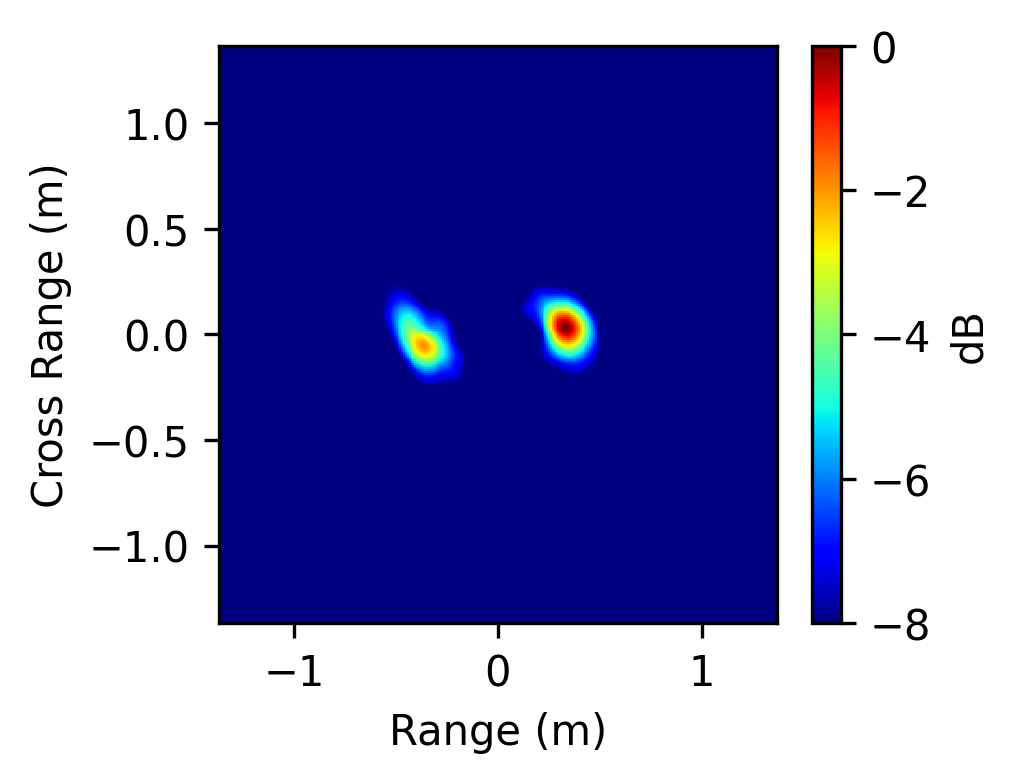} & \includegraphics[scale = 0.3]{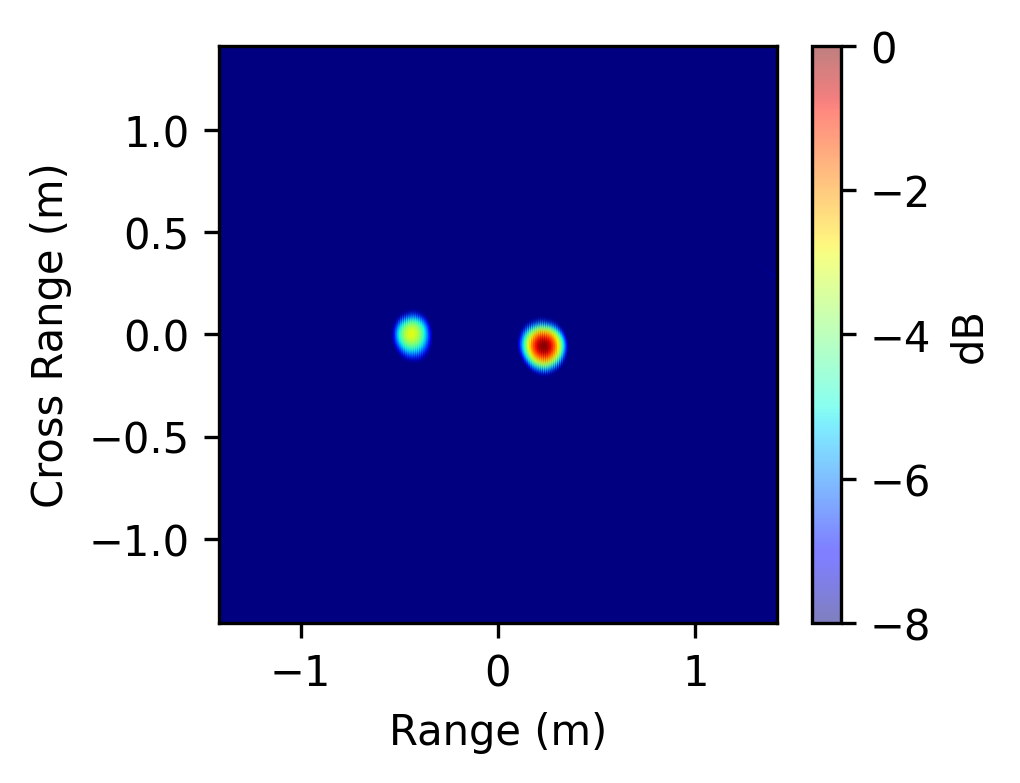} \\
    \hline
\end{tabular}

\caption{Effect of limited measurement angles on double soda cans imaging. Row 1: 90 scans (from $0^\circ$ to $90^\circ$), Row 2: 180 scans (from $0^\circ$ to $180^\circ$), and Row 3: 270 scans (from $0^\circ$ to $270^\circ$). BP fails to reconstruct the secondary object when measurements are $\leq 180^\circ$.}
\label{fig: partial_angles}
\end{figure}

\section{Applications of ATS}

\subsection{\textbf{Identification Applications: }}
Numerous studies have investigated embedding information into objects for non-intrusive content recognition. Radar-based methods like MechanoBeat \cite{MechanoBeat2020}, for instance, utilize the oscillation frequency of 3D-printed tags to embed identification information into everyday objects. Motivated by this concept, our objective in the subsequent experiments was to integrate identification information into objects such as cylindrical containers and helmets by equipping them with 3D-printed corner reflectors. We maintained a consistent number of reflectors to ensure a constant Radar Cross-Section (RCS) while varying their orientations to enhance robust detection capabilities. We utilize 3D-printed octahedral corner reflectors (20 mm radius) positioned in specific orientations to encode identification information within objects. We apply copper tapes onto the surfaces of the reflectors to further enhance their reflectivity. 
\begin{figure}
\begin{minipage}{\textwidth}
    \begin{tabular} {ccc}
        \scriptsize \textbf{Target} & \scriptsize \textbf{BP} & \scriptsize \textbf{ATS} \\
        \hline
        \includegraphics[scale = 0.019]{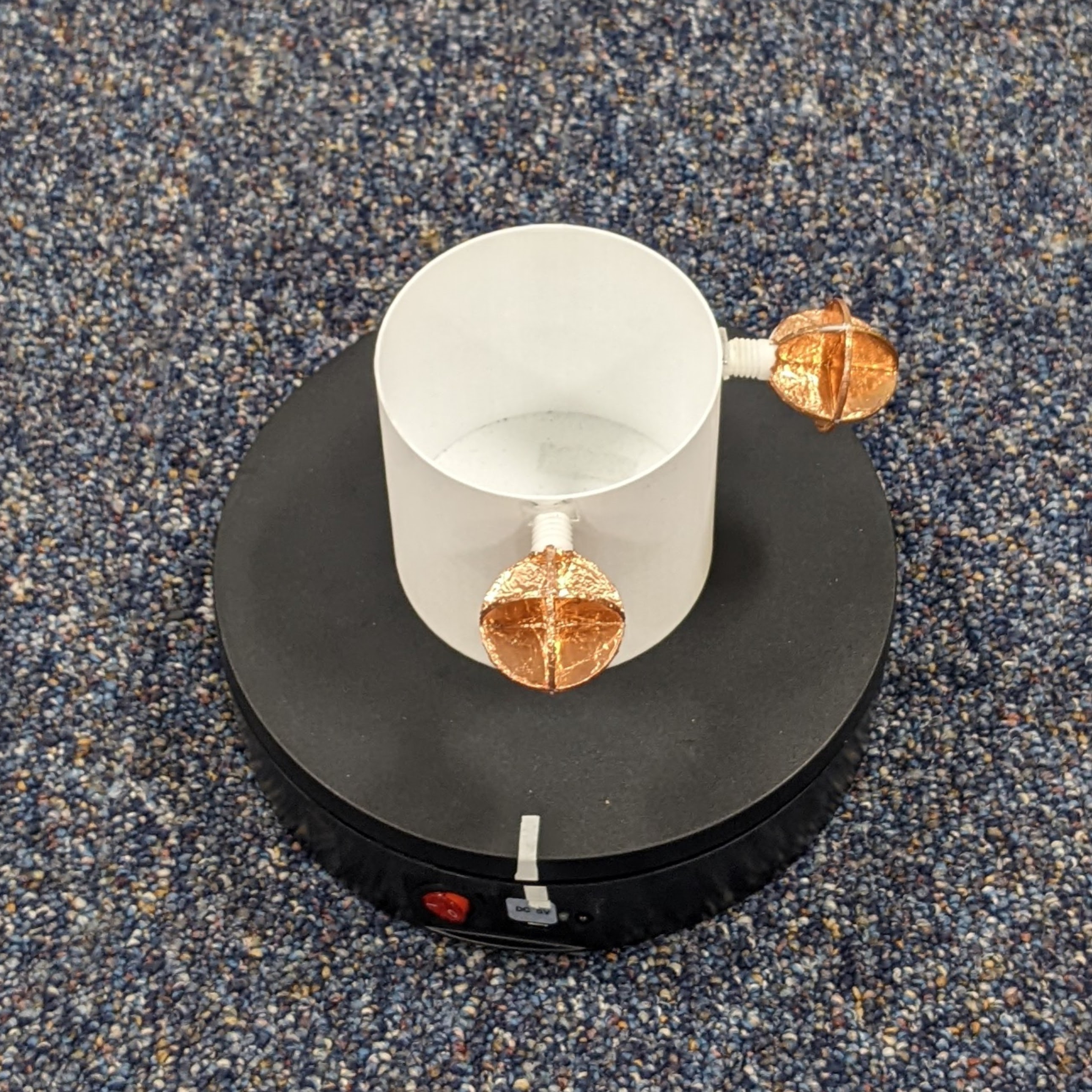} &
        \includegraphics[scale = 0.3]{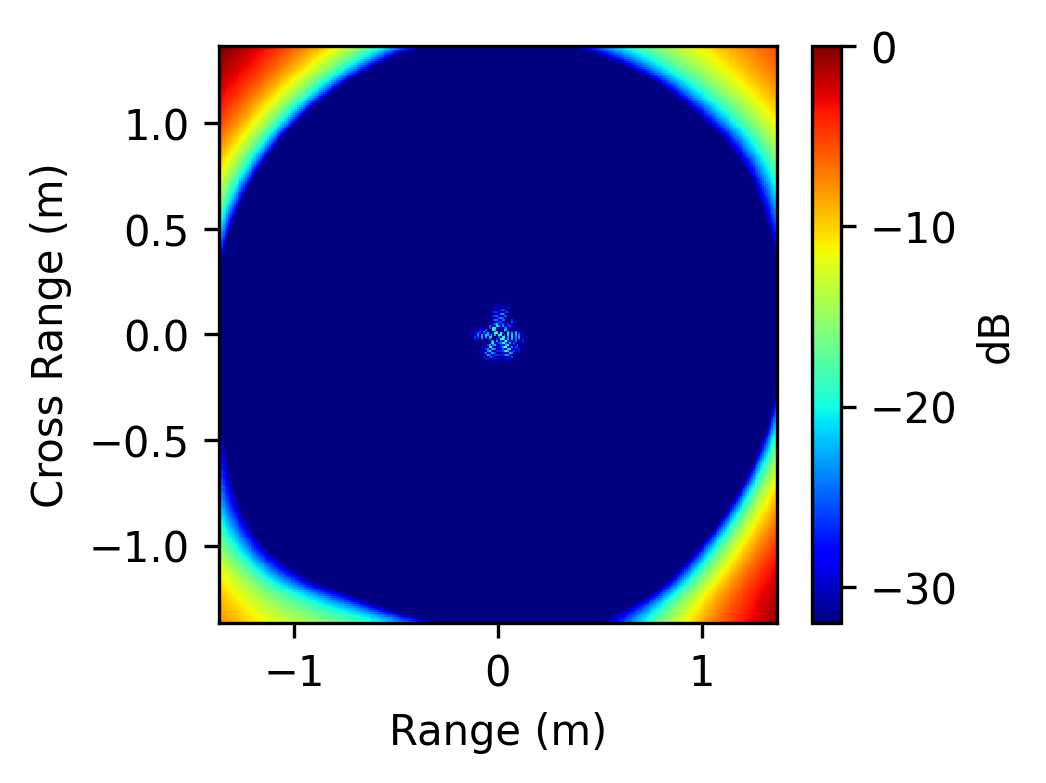} & \includegraphics[scale = 0.3]{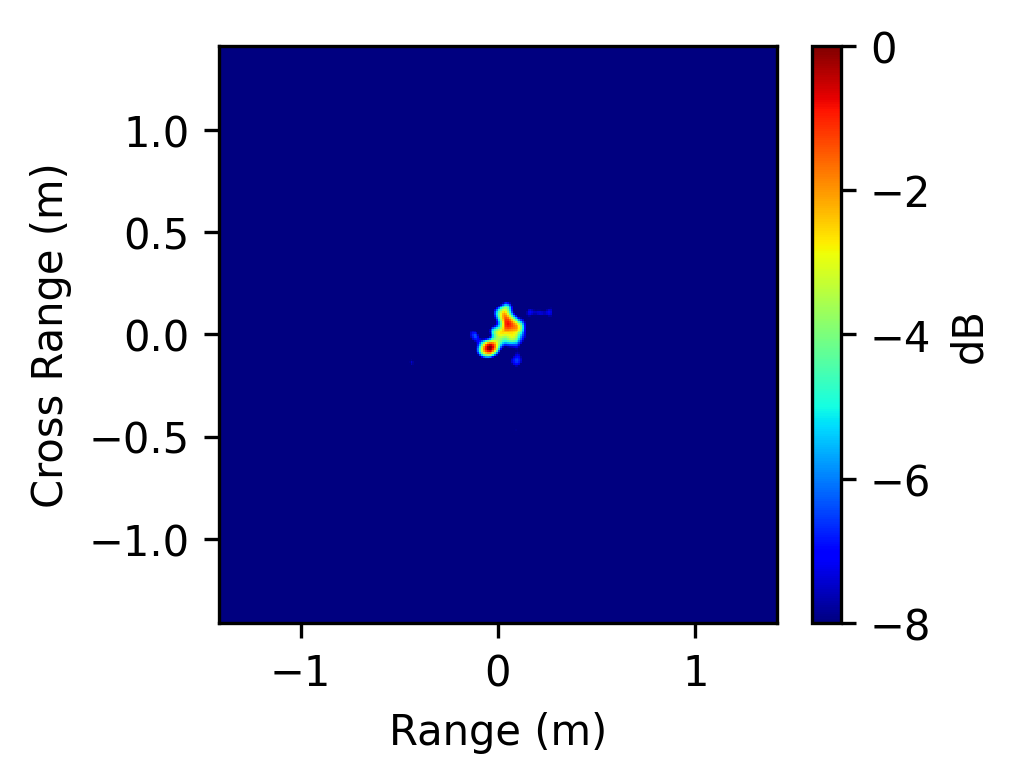}\\ 
        \hline
        \includegraphics[scale = 0.016]{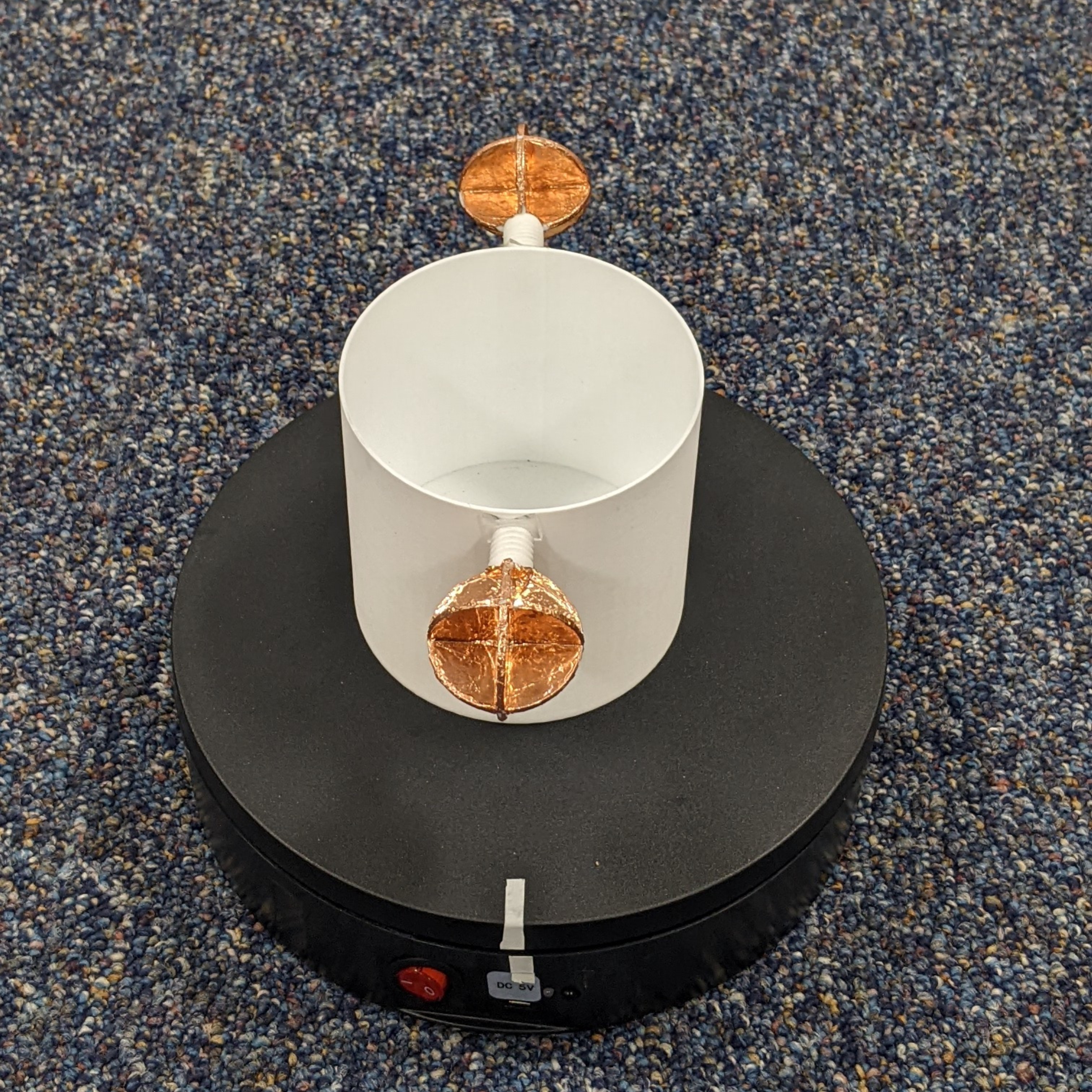} &
        \includegraphics[scale = 0.3]{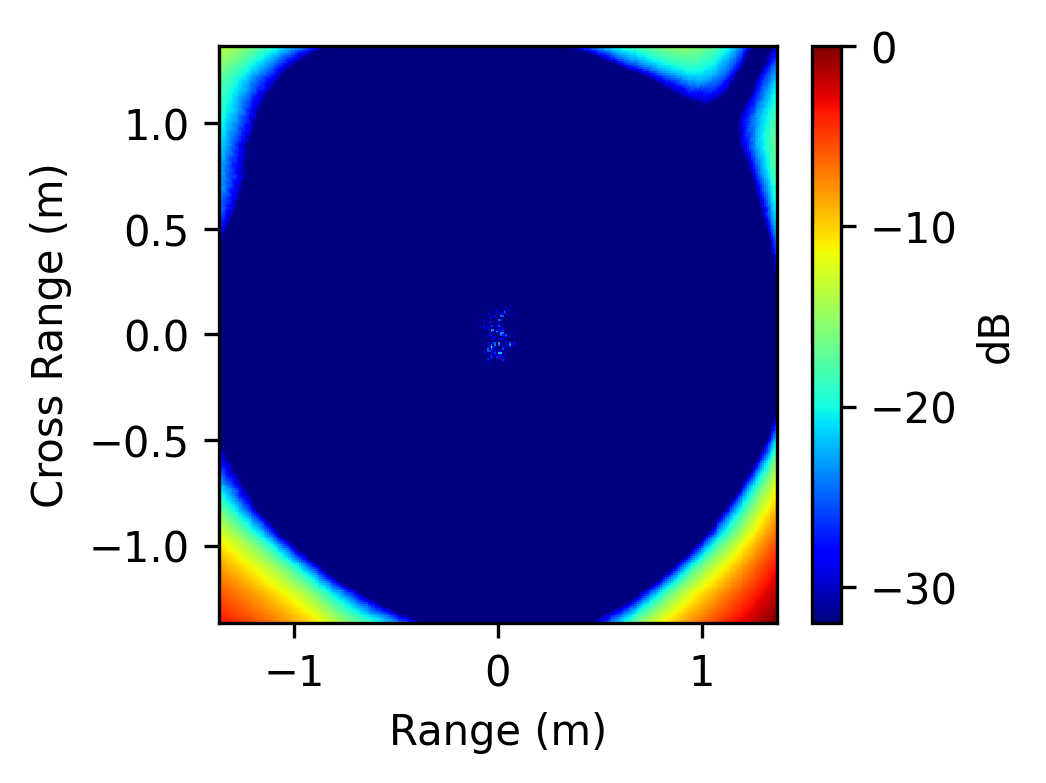} & \includegraphics[scale = 0.3]{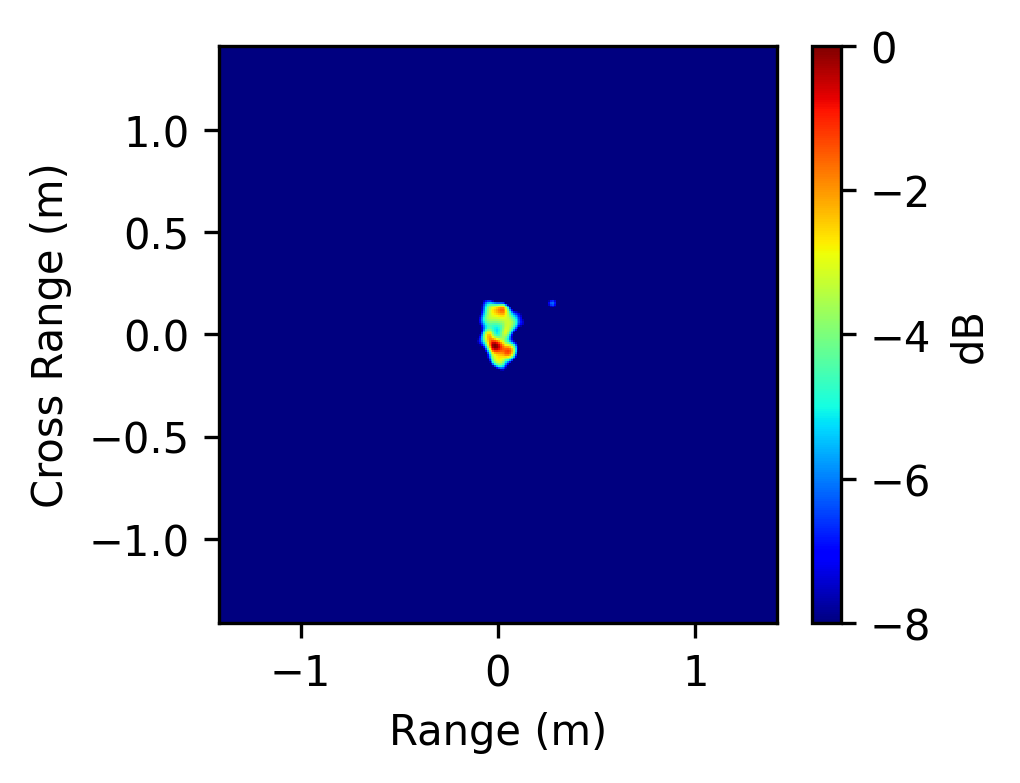}\\ 
        \hline
        \hline
        \includegraphics[scale = 0.017]{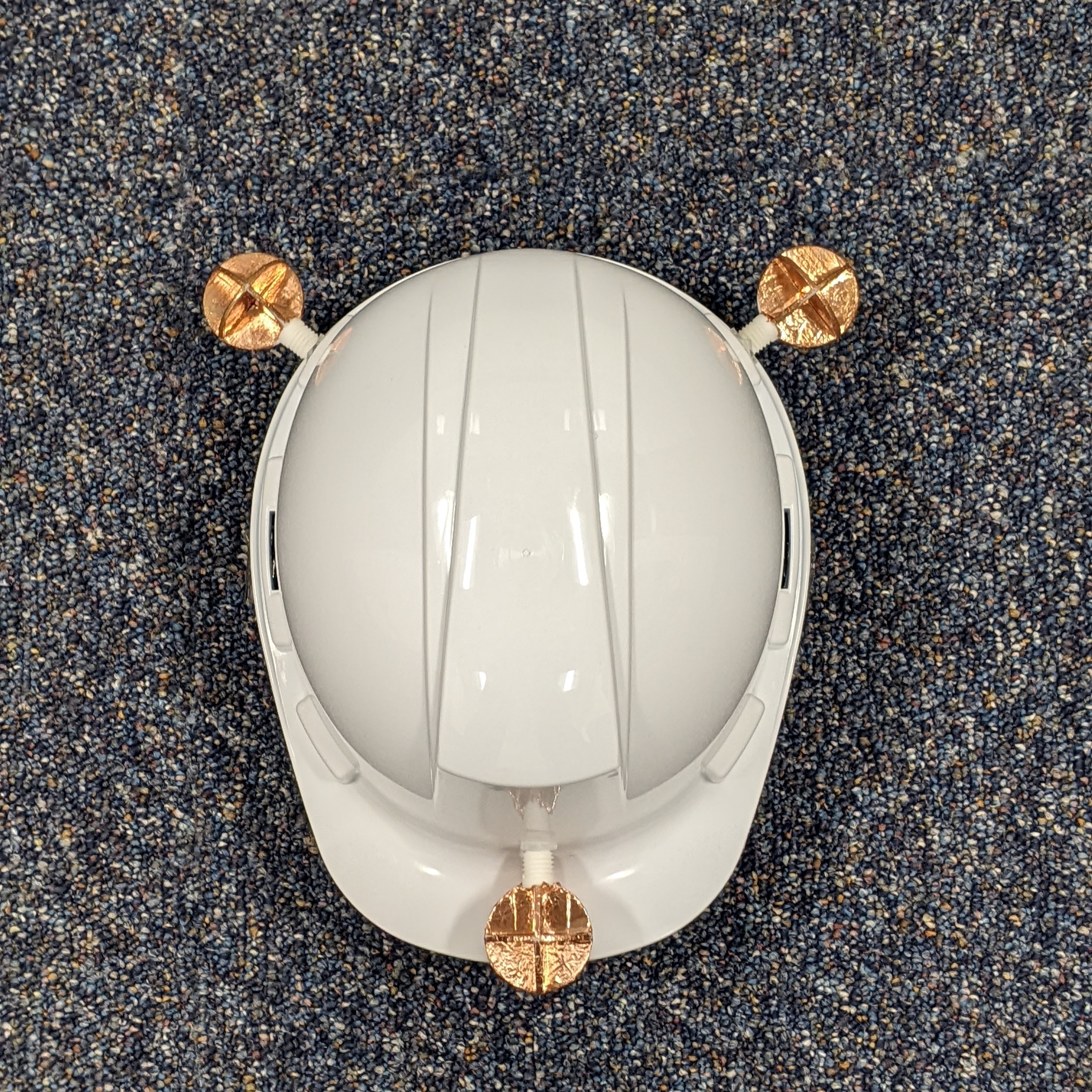} &
        \includegraphics[scale = 0.3]{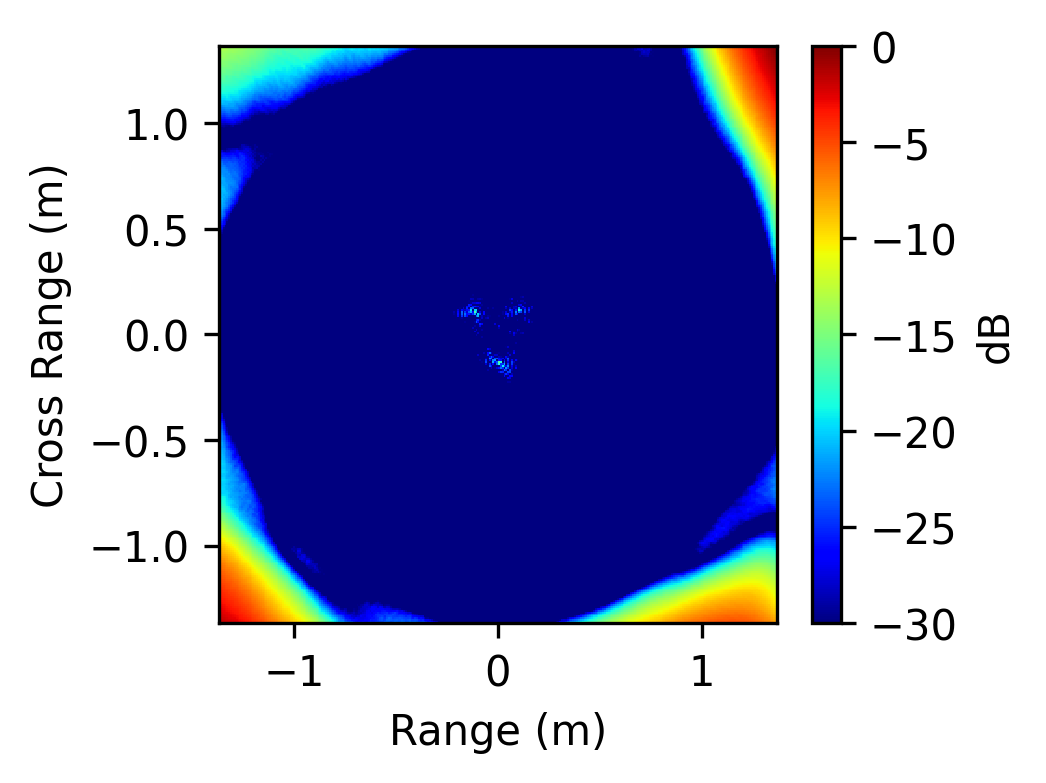} & \includegraphics[scale = 0.3]{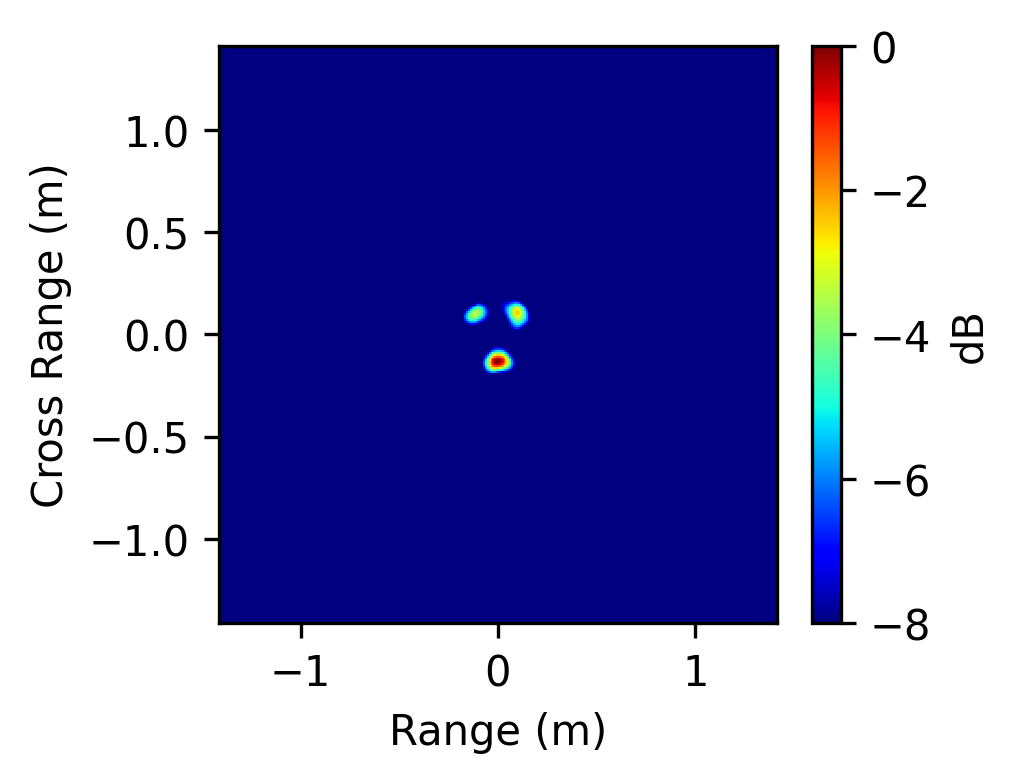}\\ 
        \hline
        \includegraphics[scale = 0.017]{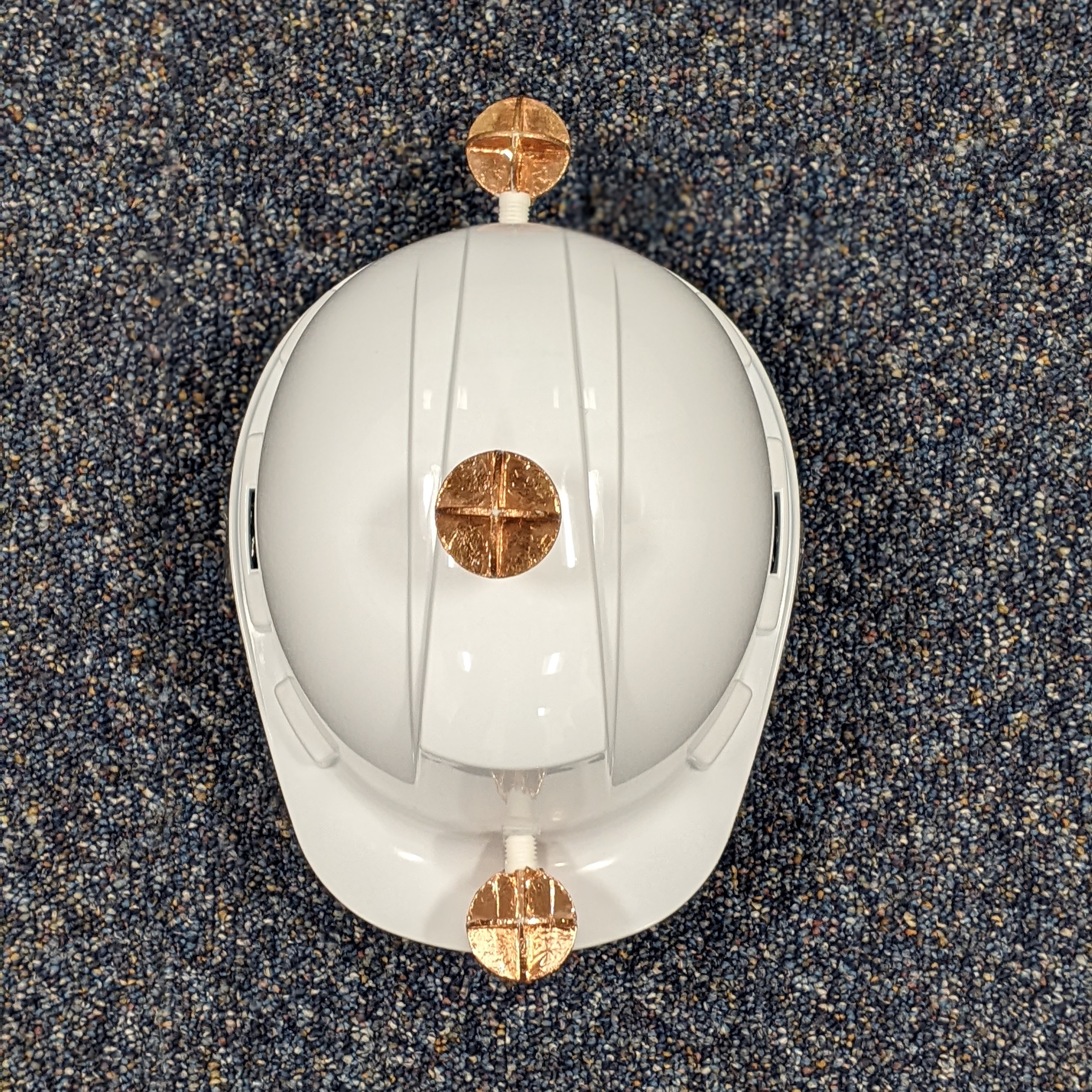} &
        \includegraphics[scale = 0.3]{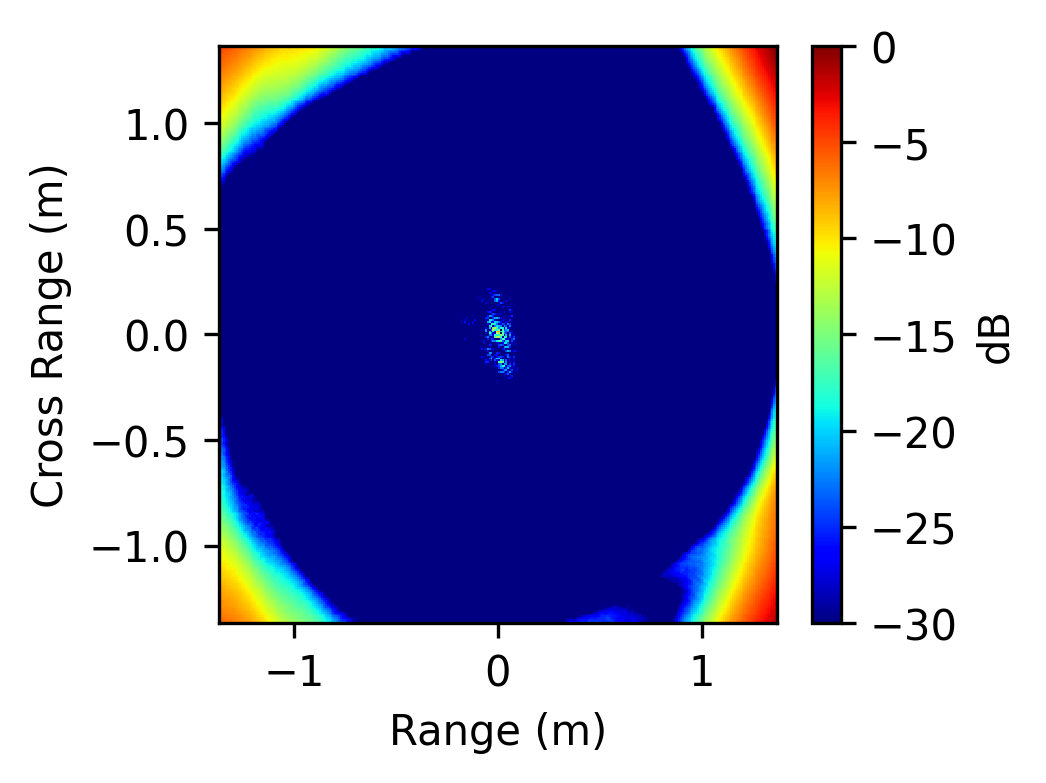} & \includegraphics[scale = 0.3]{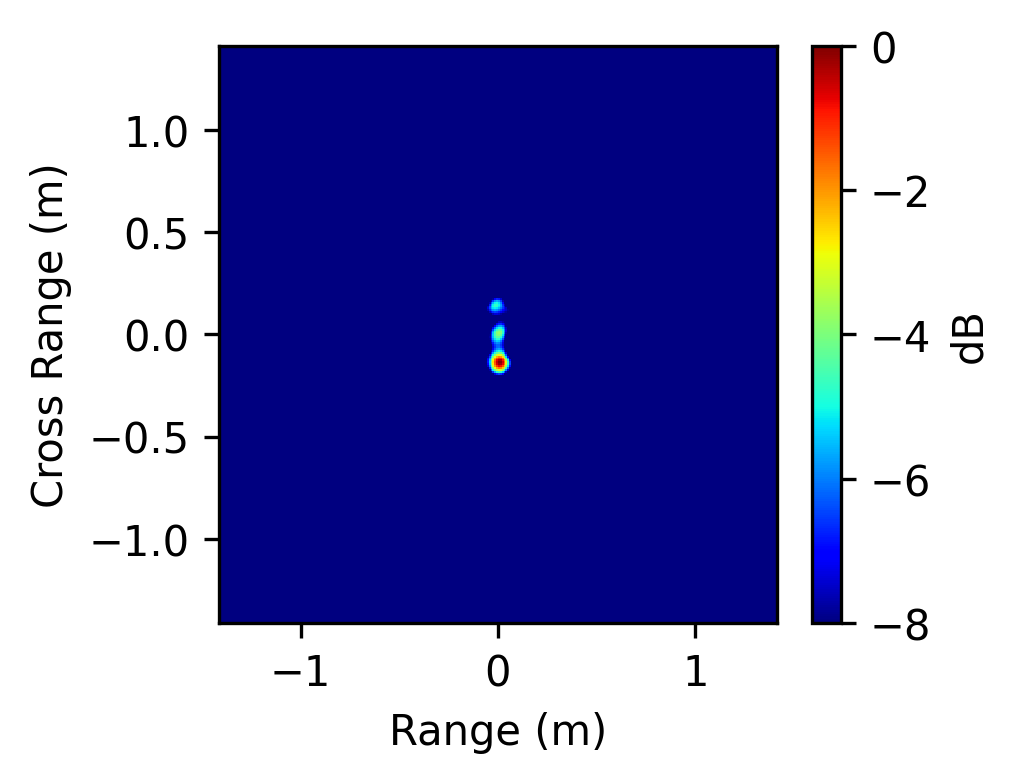}\\ 
        \hline
    \end{tabular}
\end{minipage}

\caption{Enhanced object identification using 3D printed octahedral reflectors: Top two rows depict ISAR imaging of a cylindrical container with corner reflectors placed at $90\degree$ and $180\degree$ orientations, while bottom two rows showcase ISAR imaging of a construction helmet with reflectors positioned at $120\degree$ intervals and in a linear alignment.}
\label{fig: Mug_Helmet_exp} 

\end{figure}

\subsubsection{\textbf{Plastic Container/Mug Identification}}
The top two rows of Figure \ref{fig: Mug_Helmet_exp} display imaging of a 3D-printed plastic cylindrical container having two corner reflectors positioned at $90\degree$ and $180\degree$ orientations respectively. The results show that the ATS method can reasonably reconstruct the relative positions of reflectors. In contrast, BP presents noticeable artifacts, failing to provide clear tag reconstruction even after a -30dB dynamic range adjustment.

\subsubsection{\textbf{Helmet Identification}}
In challenging environments such as construction sites and firefighting scenarios where line-of-sight may be obstructed, accurate ISAR imaging facilitates easy identification of helmets through the attached tags. Figure \ref{fig: Mug_Helmet_exp} (bottom two rows) presents imaging results of helmets fitted with 3D-printed reflector tags at varying orientations, showcasing the effectiveness of the ATS method in accurately reconstructing reflectors' positions. In contrast, BP exhibits notable artifacts and inferior tag imaging, visible only after a -30dB dynamic range adjustment.

\subsection{\textbf{Potential Imaging Applications:}}

The potential applications of ATS with radar technology span various fields. For instance, in robot-environment interaction, ATS enables robots equipped with SAR capabilities to thoroughly explore objects through scanning and reconstruction, thereby enhancing environmental understanding. In security scenarios, ATS can penetrate concealments to reconstruct the internal structures of complex objects, aiding in security clearance procedures by detecting concealed weapons. Additionally, ATS could be used for unlabelled warehouse package recognition and reconstruction, allowing for the identification of package contents without repackaging.

\section{Conclusion and Future Work}

In conclusion, we have presented an ISAR imaging algorithm that adopts an analysis-through-synthesis approach inspired by conventional NeRF techniques. By estimating scene scatterers through an implicit neural network and optimizing with a differentiable forward model incorporating radar wave propagation and reflection characteristics, our analysis-through-synthesis framework demonstrates quantitative and qualitative superiority over traditional backprojection through extensive experiments on simulated and hardware-measured data. Moreover, our approach eliminates the need for costly anechoic chambers or complex measurement testbeds, reducing both cost and computation time in the reconstruction process. Leveraging the penetration capability of our radar forward model, we achieve reconstruction in Non-Line-of-Sight scenarios such as imaging contents inside a cardboard box. 

Although ATS demonstrates superior performance over BP in scenarios with noise and limited measurement space, its drawback lies in its lack of real-time capability. Particularly in real-time applications where latency and speed are crucial, ATS is still considered slow (4.36 seconds) compared to BP (0.29 seconds) using NVIDIA RTX 3080Ti GPU. Exploring strategies such as sub-sampling or skipping may offer potential acceleration in measurement capture, warranting further investigation. Additionally, extending the framework to enable 3D volumetric reconstruction could open up possibilities for imaging applications such as robot environment interaction, security scanning, and warehouse package imaging.


\bibliography{arxiv_main}

\end{document}